\documentclass[11pt]{article}
\usepackage[]{acl}
\usepackage{times}
\usepackage{latexsym}
\usepackage[T1]{fontenc}
\usepackage[utf8]{inputenc}
\usepackage{microtype}
\usepackage{inconsolata}

\usepackage{makecell}
\usepackage{graphicx}
\usepackage{subcaption}
\usepackage{caption}
\usepackage{booktabs}
\usepackage{amsmath}
\usepackage{amssymb}
\usepackage{mathtools}
\usepackage{amsthm}
\usepackage{enumitem}
\usepackage{xspace}
\usepackage{hyperref}
\usepackage{colortbl}
\usepackage{colortbl}
\definecolor{our_green}{HTML}{E8F5E9}
\definecolor{cherry}{RGB}{222,49,99}

\newcommand{\ours}{\textsc{Pixel Linguist II}\xspace}

\title{On the Design Fundamentals of Pixel Text Representation Learning}

\author{
        \textbf{Chaohao Yuan}$^{1,2}$*\ \ 
        \textbf{Ruifeng Yuan}$^{2,3}$*\ \ 
        \textbf{Zhuoxu Huang}$^4$* \ \
        \textbf{Yu Rong}$^{2,5}$ \ \
        \textbf{Hong Cheng}$^1$ \ \
        \\
        \textbf{Hou Pong Chan}$^6$\footnotemark[2] \ \ 
        \textbf{Chenghao Xiao}$^7$\footnotemark[2] \\\\
  $^1$The Chinese University of Hong Kong
  $^2$DAMO Academy, Alibaba Group\\
  $^3$Fudan University 
  $^4$Aberystwyth University
  $^5$Hupan Lab
  $^6$University of Macau \\
  $^7$Shanghai University of Finance and Economics
  \\
}

\begin{document}
\maketitle

\renewcommand{\thefootnote}{\fnsymbol{footnote}}
\footnotetext[1]{Equal Contribution.}
\footnotetext[2]{Corresponding authors. Emails: \href{mailto:hpchan@um.edu.mo}{hpchan@um.edu.mo}, \quad\quad\quad\quad\href{mailto:xiaochenghao@sufe.edu.cn}{xiaochenghao@sufe.edu.cn}}

\begin{abstract}
Text-rich visual inputs require models that can read, retrieve, and compress language directly in pixel space, yet existing pixel-text encoders struggle with fixed resolution pretraining, visual shortcut learning, weak visual grounding, and multilingual visual text understanding. In this work, we investigate the fundamental design principles required for robust visual text representation learning. Through systematic controlled ablations, we identify four critical components: variable image resolutions and rendered font sizes provide spatial proxies for high-resolution document generalization; natural image-text pairs are indispensable for grounding and prevent text-only collapse; layout-aware rendering helps prevent pixel-level shortcuts; and a two-stage multilingual curriculum enables effective cross-lingual alignment. By integrating these principles into a scalable training recipe, we train \ours, a native-resolution vision encoder trained with on-the-fly rendering, unified contrastive grounding, and a multilingual curriculum over 280M training examples. \ours sets new state-of-the-art results on English, cross-lingual, and multilingual Visual STS and ViDoRe, while also enabling better MLLM downstream evaluation. Notably, \ours remains robust under 80\% visual token compression, showing great promise for optical context compression. Our code and resources are available at \href{https://github.com/Pixel-Linguist/Pixel-Linguist-II}{https://github.com/Pixel-Linguist/Pixel-Linguist-II}.
\end{abstract}

\section{Introduction}

Vision-language representation learning has become central to cross-modal retrieval and retrieval-augmented generation (RAG). While dual-encoder models excel on natural images and short captions, they are less suited to text-rich visual inputs, such as documents, infographics, and charts, where retrieval requires fine-grained reading, layout understanding, and document-level semantics.

Pixel-based text representation learning offers a unified alternative: rendering text directly as RGB images and encode both natural images and rendered text with a single vision encoder. Prior work has progressively shown that ViT encoders can learn language representations from pixel inputs~\cite{rust2022language}, that contrastive learning improves their discriminability~\cite{tschannen2023clippo}, and that scaled rendered-text training enables visual, topical, reasoning, and cross-lingual alignment~\cite{xiao2024pixel}. Despite this progress, robust pixel-based text representation learning still hinges on four coupled challenges: \textbf{resolution mismatch}, \textbf{visual shortcut learning}, \textbf{multimodal grounding}, and \textbf{multilingual visual text perception}. These axes determine whether a visual text encoder can move beyond synthetic rendered snippets to real-world document understanding.

Instead of simply scaling data and parameters, we ask: \textit{What are the essential design principles required to learn generalized visual text representations?} Our controlled ablations answer this question through four research questions:

\textbf{RQ1: How can computationally efficient low-resolution pretraining generalize to high-resolution documents at inference?} We find that variable natural-image resolutions and rendered font sizes act as spatial proxies, allowing small-canvas pretraining to extrapolate to dense, high-resolution documents.

\textbf{RQ2: What role does multimodal grounding play in visual text representation learning?} Natural image-text pairs remain necessary even for text-centric targets: removing them causes severe dense document retrieval degradation, while joint training grounds text semantics in real-world visual contexts.

\textbf{RQ3: How does layout diversity in text rendering affect representation quality?} Fixed fonts and plain canvases trigger pixel-level shortcut learning and near-collapse in document retrieval, showing that text rendering with diverse layouts is critical for semantic transfer.

\textbf{RQ4: What training curriculum is required for multilingual pixel-space semantics?} We find that a two-stage curriculum works best: large-scale unsupervised multilingual pretraining builds foundational capability for multilingual visual text perception, which is then activated and aligned across languages through semantic mid-training.

\begin{figure*}[h]
    \centering
    \includegraphics[width=0.92\linewidth]{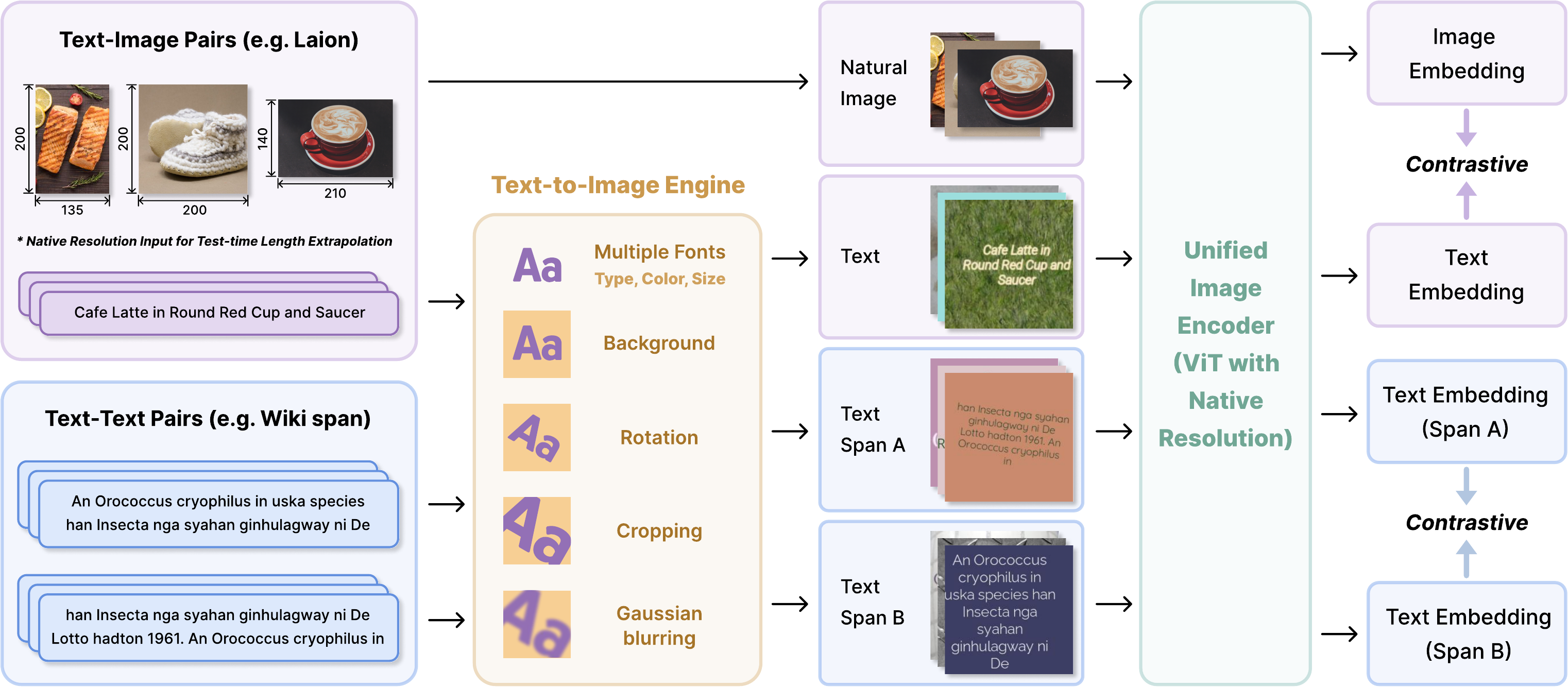}
    \caption{Overview of \ours. We construct two types of training data: natural image–text pairs and text–text pairs. Textual inputs are rendered on-the-fly into images with diverse layouts using a text-to-image rendering engine. 
    A native-resolution ViT encodes both natural images and rendered visual text within a unified pixel space. The model is trained by a contrastive learning objective.}
    \label{fig:overview}
\end{figure*}

Building on these findings, we scale the design principles into a concrete training recipe for pixel-based text representation learning. We propose \ours, a unified pixel-based vision-language representation framework for robust understanding of text in the visual modality. As illustrated in Figure~\ref{fig:overview}, \ours combines four components:

\begin{enumerate}[leftmargin=*,topsep=2pt]
\setlength{\itemsep}{2pt}
\setlength{\parskip}{0pt}
\setlength{\parsep}{0pt}
    \item \textbf{Layout-aware visual augmentation} renders text on the fly with diverse fonts, backgrounds, spatial arrangements, and visual perturbations, encouraging the model to encode semantics rather than memorizing superficial appearances.
    \item \textbf{Native-resolution encoding} adopts a Native-resolution Vision Transformer (NaViT) architecture~\cite{dehghani2023patch,bai2025qwen2} that supports variable image resolutions and aspect ratios. This design, when combined with our data pre-processing engine, enables the model to learn robust semantic extrapolation to extremely high-resolution inputs at test time.
    \item \textbf{Unified contrastive grounding} jointly trains on natural image-text pairs and rendered text-text pairs under a single contrastive objective, learning text semantics in the visual modality while grounding them in real-world visual concepts.
    \item \textbf{A multilingual training curriculum} scales learning to 280M examples through a two-stage pipeline: massive unsupervised multilingual visual text pretraining followed by high-quality semantic mid-training.
\end{enumerate}

Extensive experiments validate both the design analysis and the resulting model. \ours sets new state-of-the-art results across English, cross-lingual, and multilingual Visual STS benchmarks, achieves strong performance on the challenging ViDoRe visual document retrieval benchmark, and improves downstream performance when used as the vision encoder in multimodal large language models. Notably, visual text representations of \ours remain robust even when up to 80\% of visual tokens are compressed.

\begin{table*}[t]
\centering
\resizebox{\textwidth}{!}{
\begin{tabular}{llccccccc}
\toprule
\textbf{Pretraining Setting} & \textbf{Data Type} & \textbf{Examples} & \textbf{Batch Size} & \textbf{ArxivQA} & \textbf{InfoVQA} & \textbf{TabFQuad} & \textbf{TatDQA} & \textbf{Average} \\
\midrule
\multicolumn{8}{l}{\textit{Spatial Proxies (See Sec.~\ref{Sec: spatial proxies})}} \\
Full Components & Nat. Image + Ren. Text & 13M & 12288 & 14.99 & 56.69 & 56.05 & 23.60 & \textbf{37.83} \\
Fixed Image Size ($224\times224$) & Nat. Image + Ren. Text & 13M & 12288 & 9.61  & 46.33 & 55.23 & 22.61 & 33.45 \\
Fixed Font Size (16)     & Nat. Image + Ren. Text & 13M & 12288 & 7.83  & 48.28 & 47.89 & 19.86 & 30.97 \\
\midrule
\multicolumn{8}{l}{\textit{Multimodal Grounding \& Shortcut Removal (See Sec.~\ref{sec: multimodal grounding} \& ~\ref{sec: layout-aware augmentation})}} \\
Fixed Font + Plain Canvas & Rendered Text Only & 7M  & 12288 & 0.59  & 0.61  & 4.20  & 1.18  & 1.65  \\
\midrule
\multicolumn{8}{l}{\textit{Full Scaled-up Run (See Sec.~\ref{sec:scaled-up recipe})}} \\
Full Components (Ours)   & Nat. Image + Ren. Text & 104M & 32768 & 35.81 & 67.61 & 63.96 & 27.90 & \textbf{48.82} \\
\bottomrule
\end{tabular}
}
\caption{Deconstructing the Pixel Pretraining Pipeline. We evaluate spatial proxies and multimodal grounding using controlled, small-scale ablations on resolution-sensitive ViDoRe tasks. Performance is measured in nDCG@5.}
\label{tab:core_ablations}
\end{table*}

\section{Design Fundamentals of Pixel Text Representation Learning}

While unified visual text encoding offers a highly elegant architecture, current vision encoders remain constrained by fundamental bottlenecks: inflexible fixed-resolution processing, a lack of real-world multimodal grounding, and severe sensitivity to visual appearances. 

To overcome these limitations, we explore essential design principles required to learn generalized, real-world visual text representations.
Before running a massive-scale pretraining (Section~\ref{sec:scaled-up recipe}), we first devise a number of controlled pretraining experiments using a compact 13M-example ablation dataset consisting of natural images and rendered text, resulting in four deisgn fundamentals for generalized visual text representation learning.

\subsection{The Resolution Paradox and Spatial Proxies}
\label{Sec: spatial proxies}

A central challenge in pixel-based text encoding is the discrepancy between training and inference resolutions. Processing dense, high-resolution documents (e.g., 4K PDFs) requires encoding massive amounts of spatial information, yet pretraining is typically constrained to smaller, fixed-size canvases (e.g., $224 \times 224$ pixels) for computational efficiency. This contradiction prompts our first inquiry: 

\vspace{0.5em}
\noindent \textit{RQ1: How can a fixed-resolution canvas in training generalize to high-resolution documents in inference?}
\vspace{0.5em}

We hypothesize that computationally prohibitive high-resolution pretraining might not be strictly necessary if the network can learn the underlying concept of spatial scale through alternative means. To test this, we explore whether two factors---\textit{resolutions in natural images} and \textit{variable font sizes in rendered text}---can act as effective ``spatial proxies'' that enable the model to generalize to high-resolution documents without directly training on them. We ablate these variables during pretraining and evaluate the resulting models on resolution-sensitive document retrieval tasks (summarized in the top section of Table \ref{tab:core_ablations}).

In our controlled setup, we enforce computational efficiency while preserving variance: (1) we dynamically resize the longest side of all natural images to 224 pixels, allowing the shortest side to vary and thus preserving aspect ratio diversity without inflating compute; and (2) we render textual inputs onto a fixed $224 \times 224$ canvas while randomly sampling font sizes between 12 and 22.

Our observations reveal a clear trend: when we remove variance in natural image dimensions by resizing them to a static $224 \times 224$ resolution, average retrieval performance drops from 37.83 to 33.45. More critically, when we fix the rendered text to a uniform font size, performance degrades further to 30.97. Varying font sizes on a small canvas forces the model to encode textual features at multiple spatial frequencies. This confirms that these variances implicitly enable the model to generalize to high-resolution, dense documents during inference, bypassing the need to pretrain on large, high-resolution synthetic document canvases.

\subsection{The Necessity of Multimodal Grounding}
\label{sec: multimodal grounding}

A persistent open question in pixel-based representation learning is whether natural images are actually required if the target domain is primarily text. This leads to our second research question: 

\vspace{0.5em}
\noindent \textit{RQ2: Is synthetic rendered text alone sufficient for real-world visual text understanding? What is the role of multimodal grounding?}
\vspace{0.5em}

To answer this, we explore the boundaries of a purely synthetic visual space. We train a variant exclusively on rendered text pairs, completely removing natural image-text pairs from the pretraining corpus, mirroring approaches in previous work \cite{rust2022language,xiao2024pixel}.

We observe that while this text-only variant maintains competitive performance on simple semantic matching tasks like Visual STS, it suffers a substantial performance drop on the more complex ViDoRe benchmark.
Furthermore, when we completely isolate the model by stripping away both natural images and the spatial proxies validated in Section~\ref{Sec: spatial proxies} (i.e., training purely on rendered text with a fixed font size and static plain canvas), the degradation becomes catastrophic. As shown in the middle section of Table \ref{tab:core_ablations}, this text-only fixed template setting yields a near-zero average document retrieval score of just 1.65.

These findings suggest that training an encoder in an isolated, synthetic pixel space does not offer a valid solution for real-world document understanding. Natural image-text pairs act as a fundamental regularizer that grounds synthetic textual semantics in real-world visual contexts and exposes the model to the heterogeneous layout structures necessary for document processing. Consequently, joint multimodal training is strictly required to prevent representation collapse.

Later in Figure~\ref{fig:ablation_study_with_without_nautral_image}, we conduct the same ablation using our full-scale data, showing similar findings. This suggests that the necessity of multimodal grounding can not be bypassed by data scale alone. 

\subsection{Mitigating Shortcut Learning via Layout Augmentation}
\label{sec: layout-aware augmentation}

A related bottleneck in pixel-based text encoding is shortcut memorization. When text is rendered using fixed visual templates (e.g., uniform fonts and plain backgrounds), vision encoders naturally gravitate toward overfitting to superficial visual attributes. Returning to the theme of visual variation established in RQ1, we ask:

\vspace{0.5em}
\noindent \textit{RQ3: How does layout diversity affect representation quality?}
\vspace{0.5em}

The vulnerability caused by this sensitivity is evident in our core ablations (Table \ref{tab:core_ablations}). As previously noted, stripping away layout diversity by fixing the font size significantly degrades retrieval performance from 37.83 to 30.97. Furthermore, removing both font variance and background diversity (the ``Plain Canvas'' setting) leads to the catastrophic collapse observed in Section~\ref{sec: multimodal grounding}.

These ablations show that models must be forced to abstract away from pixel-level shortcuts. Thus, we introduce layout-aware visual augmentation. During on-the-fly text rendering, we inject structured layout diversity on fonts and backgrounds, detailed in Section~\ref{sec: full layout diversity engine}.
By ensuring the model never encounters the exact same visual instantiation of a text twice, we effectively force the encoder to prioritize semantic structure over visual shortcuts.

\subsection{The Data Curriculum: Activating Multilingual Pixels}
\label{sec: multilingual curriculum}
Having established the fundamental rendering and grounding principles, our final exploration focuses on the training trajectory itself. While high-quality curated semantic pairs are sufficient to train an English-only visual text encoder, extending this capability to a global multilingual pixel space introduces a severe bottleneck. Character sets such as Arabic, Chinese, and Korean exhibit vastly different visual and spatial structures compared to the Latin alphabet. Thus, we ask: 

\noindent \textit{RQ4: What curriculum is required to inject foundation capabilities for multilingual visual text understanding?}
\vspace{0.5em}

We investigate whether a model can learn cross-lingual semantic alignment purely from high-quality curated pairs, or if it fundamentally requires prior perceptual knowledge of these scripts. We evaluate two curriculum settings on the cross-lingual and multilingual subsets of Visual STS: one model trained \textit{exclusively} on high-quality semantic pairs (Mid-Train Only), and another that first undergoes large-scale unsupervised contrastive pretraining on highly dense multilingual Wikipedia articles before semantic tuning (Pretrain + Mid-Train).

As shown in Table \ref{tab:curriculum_ablation}, relying solely on curated cross-lingual semantic pairs yields a performance ceiling. However, introducing a foundational stage of unsupervised pretraining provides a consistent boost of $\sim$3.3 to 3.5 absolute points across diverse languages. This establishes a core data curriculum principle: massive unsupervised multilingual visual text pretraining acts as a foundational training phase to inject multilingual visual text perceptual capability,
which is subsequently \textit{activated} and refined during the semantic mid-training phase.

\begin{table}[t]
\centering
\resizebox{\columnwidth}{!}{
\begin{tabular}{lcc}
\toprule
\textbf{Curriculum Setting} & \textbf{Cross-lingual} & \textbf{Multilingual} \\
\midrule
Curated Semantic Pairs Only & 53.83 & 61.79 \\
\textbf{Unsup. Pretraining + Curated Pairs} & \textbf{57.16} & \textbf{65.27} \\
\bottomrule
\end{tabular}
}
\caption{Data Curriculum Ablation on Visual STS (Spearman Correlation).}
\label{tab:curriculum_ablation}
\vspace{-10pt}
\end{table}

\section{Instantiating the Recipe: \ours}
\label{sec:scaled-up recipe}

Having established the fundamental design principles for optical text representation, we instantiate our methodology at scale. Our resulting model, \ours, integrates native-resolution processing, multimodal grounding, layout-aware augmentation, and a strict data curriculum into a unified vision-only architecture.

\subsection{Native-Resolution Encoding Architecture}

To fully leverage the spatial proxies identified in Section~\ref{Sec: spatial proxies} (i.e., variable image resolutions and font sizes), the vision backbone must natively support arbitrary aspect ratios and resolutions without lossy resizing. We adopt a Native-resolution Vision Transformer (NaViT) architecture, initializing the ViT parameters from Qwen2.5-VL's ViT. By processing a variable number of visual tokens rather than relying on fixed-grid interpolation, the encoder preserves the fine-grained structural integrity of dense document layouts and small text. Additionally, a $2\times2$ pooling layer is applied to compress adjacent visual tokens, balancing semantic capability and encoding efficiency during modeling.

\subsection{Layout-Aware Rendering Engine}
\label{sec: full layout diversity engine}

To implement the augmentation requirements established in Section~\ref{sec: layout-aware augmentation}, we develop an on-the-fly text-to-image rendering engine. Textual inputs are rendered dynamically at each epoch, ensuring the model never ground text semantics in visual shortcuts. We sample from 393 unique fonts (Table~\ref{tab:font_lib}) across languages, and stochastically apply background variations, including brightness jittering, Gaussian blur, and over 5,000 distinct textured backgrounds from the Describable Textures Dataset (DTD) \cite{cimpoi2014describing}. In Figure~\ref{fig:examples}, we provide examples of multilingual texts rendered using our rendering engine.

\subsection{Unified Contrastive Grounding}
\label{sec: method: multimodal grounding}

Based on the multimodal grounding requirements established in Section~\ref{sec: multimodal grounding}, we design a data recipe for unified contrastive grounding, incorporating both text-text pairs and text-image pairs. For \textbf{text-text pairs}, we leverage both high-quality multilingual text pretraining copus (used in first-stage training) and high-quality multilingual text pair datasets (used in second-stage training), detailed in the next subsection (Section~\ref{sec: method: multilingual curriculum}). For \textbf{image-text pairs} we sample 26M natural image-text pairs sourced from LAION-2B \cite{schuhmann2022laion} to maintain real-world grounding. This design serves as a regularizer to maintain the model's world knowledge, preventing the model from learning textual semantics purely from shapes.

\subsection{The Scaled Multilingual Curriculum}
\label{sec: method: multilingual curriculum}

\begin{table*}[ht]
\centering
\resizebox{\textwidth}{!}{%
\begin{tabular}{lcccccccc}
\toprule
\textbf{Model name} & \textbf{v-STS12} & \textbf{v-STS13} & \textbf{v-STS14} & \textbf{v-STS15} & \textbf{v-STS16} & \textbf{v-STS17} & \textbf{v-STS-b} & \textbf{Avg.} \\
\midrule
google/siglip-base-patch16-224 & 63.19 & 55.40 & 57.99 & 73.07 & 67.79 & 77.78 & 54.50 & 64.25 \\
openai/clip-vit-large-patch14 & 53.89 & 66.78 & 55.98 & 72.03 & 70.49 & 75.26 & 56.74 & 64.45 \\
laion/CLIP-ViT-H-14-laion2B-s32B-b79K & 57.00 & 62.25 & 58.62 & 74.40 & 70.57 & 76.69 & 58.99 & 65.50 \\
openai/clip-vit-base-patch16 & 63.82 & 63.26 & 56.99 & 73.32 & 68.91 & 78.18 & 57.93 & 66.06 \\
google/siglip-so400m-patch14-384 & 61.90 & 62.95 & 60.58 & 76.17 & 73.48 & 78.41 & 62.63 & 68.02 \\
EVA02-CLIP-bigE-14 & 62.24 & 62.36 & 62.17 & 77.41 & 73.63 & 80.96 & 62.85 & 68.80 \\
google/siglip-large-patch16-384 & 66.30 & 62.08 & 61.66 & 77.11 & 73.27 & 79.58 & 66.59 & 69.51 \\
laion/CLIP-ViT-bigG-14-laion2B-39B-b160k & 62.81 & 68.16 & 65.50 & 78.67 & 74.89 & 79.97 & 66.54 & 70.93 \\
EVA02-CLIP-bigE-14-plus & 63.36 & 68.00 & 66.38 & 79.45 & 75.26 & 82.87 & 68.59 & 71.99 \\
\midrule
\multicolumn{9}{c}{\textbf{\textit{Backbone}}} \\
Qwen2.5-VIT & 47.50 & 36.49 & 30.95 & 54.69 & 53.71 & 63.87 & 38.63 & 46.55\\
\multicolumn{9}{c}{\textbf{\textit{Ours}}} \\
\ours (mid-training only) & 65.78 & 70.00 & 67.76 & 82.39 & 76.99 & 84.83 & 75.30 & 74.72\\ 
\ours (mid-training + finetuning) & 76.60 & 75.94 & 75.07 & 85.17 & 79.65 & 85.25 & 80.93 & 79.80\\
\bottomrule
\end{tabular}%
}
\caption{\ours Performance on Visual STS Tasks (English-only) \cite{xiao2024pixel,xiao2025mieb}, which renders traditional STS tasks in NLP as image-only tasks. This task assesses the ability of vision encoders on textual semantic understanding on text-rich images.}
\label{tab:sts_benchmarks}
\vspace{-5pt}
\end{table*}

To faciliate high multilingual visual text understanding capability, we instanstiate the two-stage training recipe established in Section~\ref{sec: multilingual curriculum} consisting of two text dataset types: (1) multilingual pretraining corpus. (2) high-quality text pairs. 

For \textbf{multilingual pretraining corpus} (referred to as \textit{Text Corpus 1}), we leverage multilingual Wikipedia pretraining corpus of 62M documents. For each document, we randomly crop 25\% to 50\% each document twice to serve as unsupervised positive pairs \cite{izacard2021unsupervised}. For \textbf{high-quality semantic text pairs} (referred to as \textit{Text Corpus 2}), we curate 26M pairs from high-quality datasets used for text embedding model training.

Combining our multimodal and multilingual dataset recipes, the training curriculum is divided into two phases:
\begin{itemize}[leftmargin=*,topsep=2pt]
\setlength{\itemsep}{2pt}
\setlength{\parskip}{0pt}
\setlength{\parsep}{0pt}
    \item \textbf{Stage 1: Foundational Pretraining} combines \textit{Text Corpus 1} (62M examples) and \textit{image-text pairs} (26M examples) 
    \item \textbf{Stage 2: Semantic Mid-Training:} combines \textit{Text Corpus 2} (26M examples) and \textit{image-text pairs} (26M examples) 
\end{itemize}

Each stage is run for 2 epochs, resulting in a total examples seen of 280 millions. 

\subsection{Training Implementation Details}

We implement distributed data parallel (DDP) training with DeepSpeed ZeRO 2. Representations are all-gathered across all GPUs and nodes to compute the InfoNCE loss \cite{oord2018representation}, after which gradients are backpropagated to each GPU. We use a global batch size of 32,768 across 64 GPUs, with a per-device batch size of 512, and set the temperature to 0.03.

\section{Main Results}

\begin{figure}[h]
    \centering
    \includegraphics[width=\linewidth]{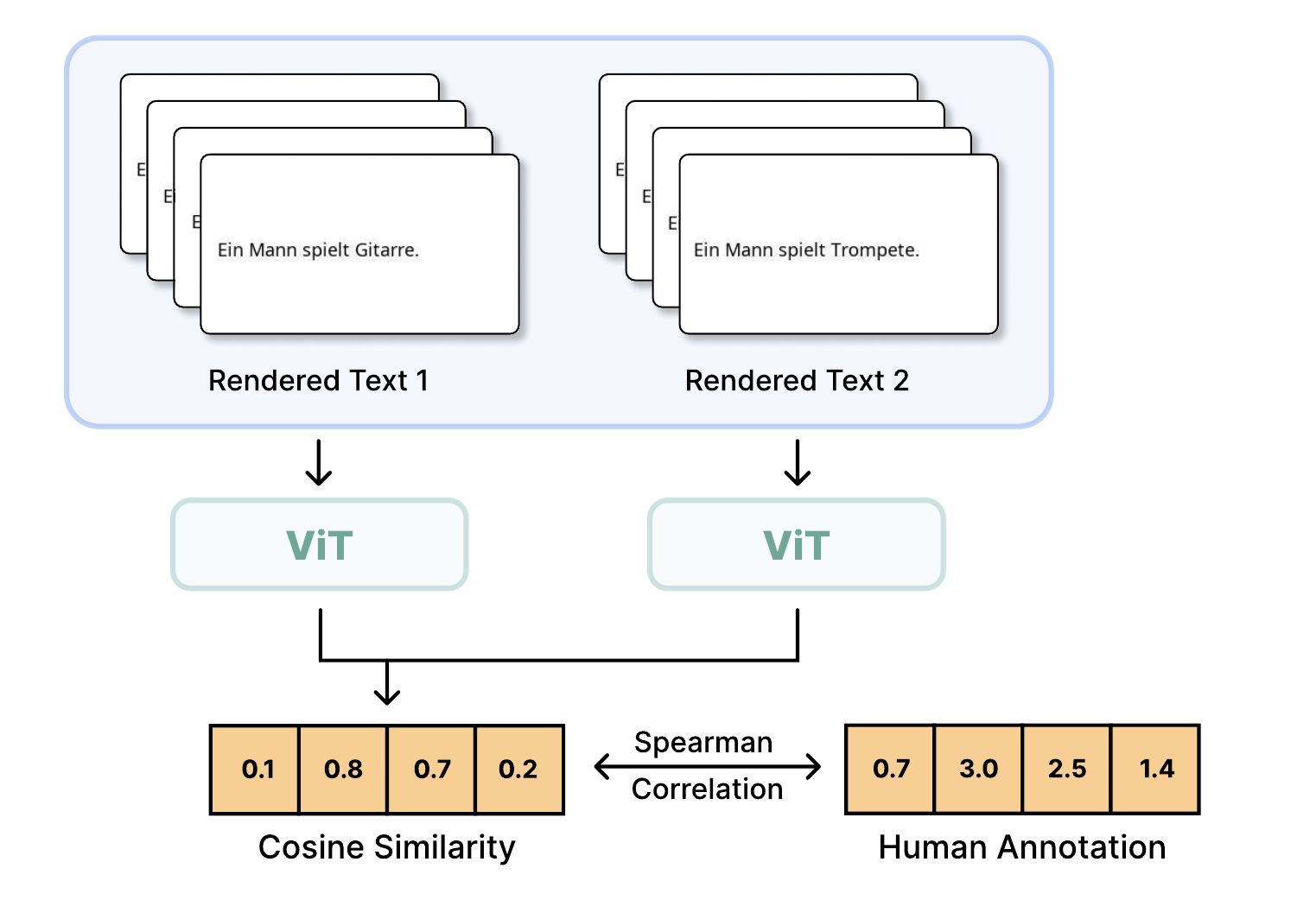}
    \caption{Visual STS task. Text pairs are rendered as images for models to quantify their semantic similarity.}
    \label{fig:visual-sts}
\end{figure}

\begin{figure}[h]
    \centering
    \includegraphics[width=\linewidth]{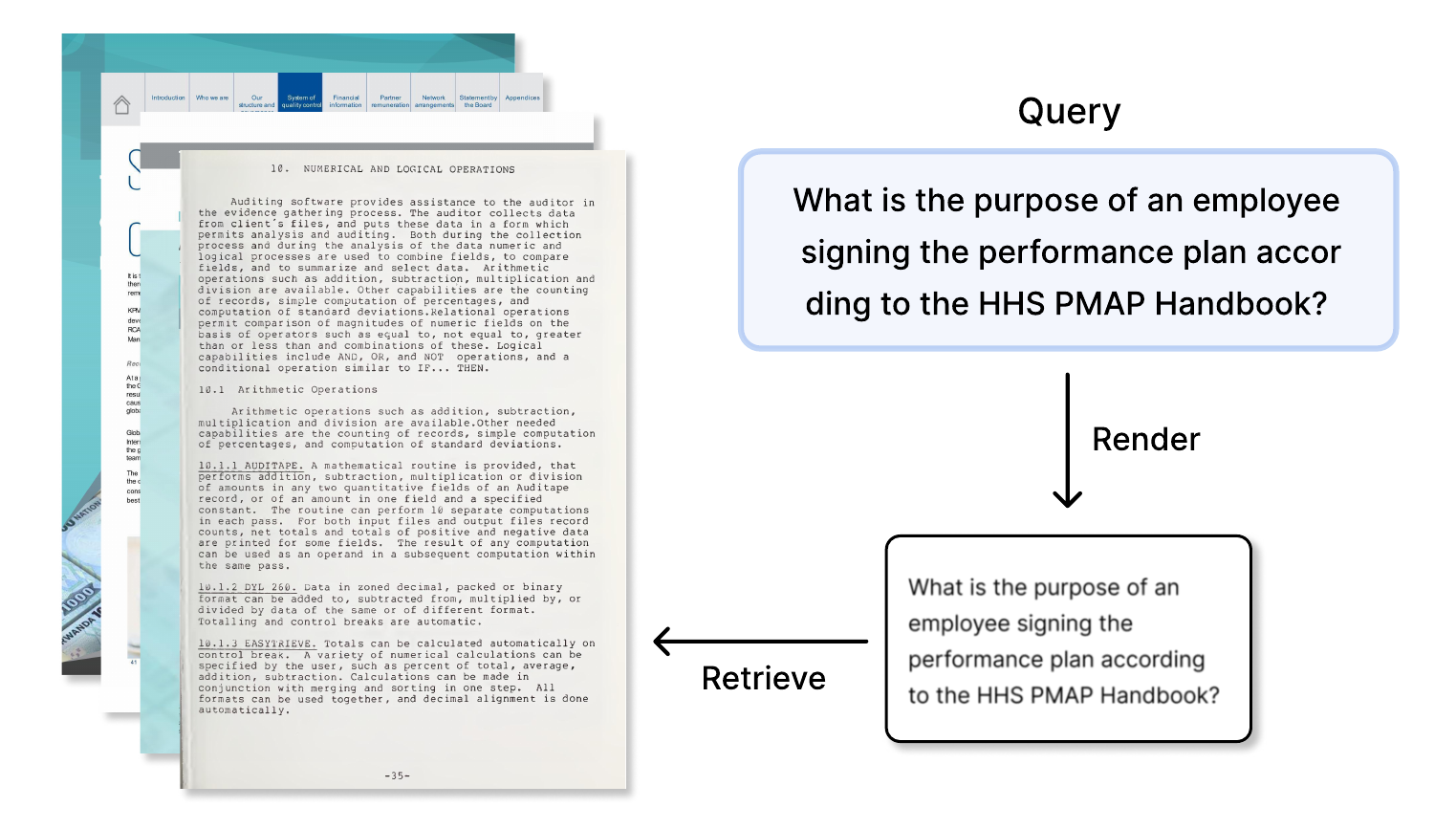}
    \caption{Visual document retrieval (VDR) task. Note that \ours processes VDR tasks in an unified way, i.e., the text queries are also first rendered as images and processed visually, as opposed to CLIP-style models we benchmark against.}
    \label{fig:VDR_illustration}
\end{figure}

\begin{table*}[ht]
\centering
\resizebox{\textwidth}{!}{%
\begin{tabular}{lccccccc}
\toprule
\textbf{Model name} & \textbf{DocVQA} & \textbf{InfoVQA} & \textbf{Sft Proj.} & \textbf{AI} & \textbf{Tabfquad} & \textbf{Tatdqa} & \textbf{Avg.} \\
\midrule
\multicolumn{8}{c}{\textbf{\textit{Baselines}}} \\
openai/clip-vit-base-patch16 & 14.60 & 51.70 & 7.13 & 22.86 & 17.61 & 4.71 & 19.77 \\
google/siglip-base-patch16-224 & 16.04 & 46.11 & 3.71 & 25.27 & 29.04 & 5.08 & 20.87 \\
EVA02-CLIP-bigE-14 & 16.35 & 54.80 & 10.14 & 33.53 & 28.80 & 7.09 & 25.12 \\
EVA02-CLIP-bigE-14-plus & 16.84 & 55.19 & 12.76 & 34.57 & 30.36 & 7.52 & 26.21 \\
openai/clip-vit-large-patch14 & 16.69 & 62.44 & 17.05 & 38.25 & 30.95 & 11.00 & 29.40 \\
laion/CLIP-ViT-L-14-DataComp.XL-s13B-b90K & 19.68 & 55.61 & 16.19 & 47.20 & 30.70 & 15.27 & 30.78 \\
google/siglip-large-patch16-256 & 22.39 & 54.09 & 9.13 & 43.40 & 49.81 & 12.38 & 31.87 \\
laion/CLIP-ViT-bigG-14-laion2B-39B-b160k & 20.44 & 60.90 & 25.02 & 55.42 & 35.02 & 16.21 & 35.50 \\
google/siglip-so400m-patch14-384 & 31.28 & 69.73 & 25.04 & 67.78 & 60.29 & 27.52 & 46.94 \\
\midrule
\multicolumn{8}{c}{\textbf{\textit{Ablation}}} \\
google/siglip-so400m-patch14-384 & 31.28 & 69.73 & 25.04 & 67.78 & 60.29 & 27.52 & 46.94 \\
google/siglip-so400m-patch14-384 (vision-only eval.) & 12.05 & 34.70 & 8.96 & 34.54 & 34.67 & 11.81 & 22.79 \\
$\Delta$ Performance Difference & 19.23\textcolor{cherry}{\rlap{$\downarrow$}} & 35.03\textcolor{cherry}{\rlap{$\downarrow$}} & 16.08\textcolor{cherry}{\rlap{$\downarrow$}} & 33.24\textcolor{cherry}{\rlap{$\downarrow$}} & 25.62\textcolor{cherry}{\rlap{$\downarrow$}} & 15.71\textcolor{cherry}{\rlap{$\downarrow$}} & 24.15\textcolor{cherry}{\rlap{$\downarrow$}} \\
\midrule
\multicolumn{8}{c}{\textbf{\textit{Backbone}}} \\
Qwen2.5-VIT & 0.94 & 0.74 & 0.93 & 0.89 & 7.15 & 2.2 & 2.14 \\
\multicolumn{8}{c}{\textbf{\textit{Ours}}} \\
\ours (mid-training only) & 20.46 & 67.61 & 37.20 & 75.09 & 63.96 & 27.90 & 48.70 \\
\ours (mid-training + finetuned) & 20.91 & 69.37 & 41.60 & 72.94 & 71.46 & 29.36 & 50.94 \\
\bottomrule
\end{tabular}%
}
\caption{Encoder performance of \ours on Visual Document Retrieval (VDR) Tasks using ViDoRe subsets~\cite{faysse2025colpali} in MIEB-lite benchmark~\cite{xiao2025mieb}, compared with SOTA baseline encoder models.}
\label{tab:VDR}
\end{table*}

\begin{table*}[ht]
\centering
\resizebox{\textwidth}{!}{%
\begin{tabular}{lccccccccccc}
\toprule
\textbf{Model name} & \textbf{ko-ko} & \textbf{ar-ar} & \textbf{en-ar} & \textbf{en-de} & \textbf{en-tr} & \textbf{es-en} & \textbf{es-es} & \textbf{fr-en} & \textbf{it-en} & \textbf{nl-en} & \textbf{Avg.} \\
\midrule
EVA02-CLIP-bigE-14-plus & 11.36 & 31.51 & 10.71 & 24.33 & -10.05 & 20.18 & 59.20 & 36.12 & 28.60 & 33.18 & 24.52 \\
EVA02-CLIP-bigE-14 & 10.97 & 29.99 & 13.49 & 22.76 & 6.39 & 29.03 & 57.16 & 36.66 & 33.43 & 26.16 & 26.60 \\
google/siglip-base-patch16-224 & 21.00 & 25.03 & 14.36 & 31.20 & 24.80 & 29.32 & 69.85 & 35.70 & 27.46 & 28.98 & 30.77 \\
laion/CLIP-ViT-bigG-14-laion2B-39B-b160k & 14.38 & 32.39 & 12.21 & 36.74 & 14.99 & 30.44 & 69.77 & 39.77 & 36.44 & 34.83 & 32.20 \\
laion/CLIP-ViT-H-14-laion2B-s32B-b79K & 19.39 & 33.39 & 19.49 & 43.78 & 16.68 & 27.99 & 62.58 & 39.32 & 28.59 & 37.33 & 32.85 \\
openai/clip-vit-base-patch16 & 10.54 & 36.25 & 13.13 & 41.57 & 35.42 & 24.63 & 62.95 & 38.72 & 31.40 & 38.63 & 33.32 \\
laion/CLIP-ViT-L-14-DataComp.XL-s13B-b90K & 14.28 & 36.47 & 12.75 & 43.10 & 19.70 & 37.37 & 71.62 & 36.88 & 30.78 & 30.76 & 33.37 \\
openai/clip-vit-large-patch14 & 11.07 & 39.12 & 18.95 & 45.71 & 39.70 & 36.76 & 70.11 & 44.06 & 40.17 & 41.63 & 38.73 \\
google/siglip-so400m-patch14-384 & 13.65 & 45.76 & 11.22 & 46.07 & 30.62 & 40.08 & 73.62 & 46.36 & 36.45 & 44.95 & 38.88 \\
\midrule
\multicolumn{12}{l}{\textbf{\textit{Backbone}}} \\
Qwen2.5-ViT & 51.34 & 52.45 & 22.07 & 24.77 & 22.58 & 16.71 & 65.44 & 32.05 & 26.00 & 26.43 & 33.98\\
\multicolumn{12}{c}
{\textbf{\textit{Ours}}}\\
\ours (mid-training) & 51.13 & 50.96 & 2.09 & 66.00 & 54.59 & 55.51 & 70.66 & 63.00 & 61.91 & 62.46 & 53.83  \\ 
\ours (pre-training + mid-training) & 57.51 & 50.13 & 1.44 & 67.4 & 55.86 & 61.42 & 75.82 & 67.73 & 67.88 & 66.36 & 57.16 \\
\bottomrule
\end{tabular}%
}
\caption{Encoder Performance of \ours on Visual STS Tasks (Cross-lingual Tasks)}
\label{tab:sts_crosslingual_benchmarks}
\end{table*}

\paragraph{Overview}

We evaluate \ours on Visual Semantic Textual Similarity (Visual STS)~\cite{xiao2024pixel,xiao2025mieb} and Visual Document Retrieval (VDR)~\cite{faysse2025colpali}, comparing against 18 competitive baselines, including CLIP~\cite{radford2021learning}, OpenCLIP~\cite{ilharco2021openclip}, DataComp-CLIP~\cite{gadre2023datacomp}, SigLIP~\cite{zhai2023sigmoid}, and EVA-CLIP~\cite{sun2023eva}. Tables~\ref{tab:sts_benchmarks} and~\ref{tab:VDR} compare strongest baselines on Visual STS and VDR, respectively (See Tables~\ref{appendix:tab:sts_benchmarks} and~\ref{appendix:tab:VDR} for all model results). We illustrate the Visual STS and Visual Document Retrieval task settings in Figure~\ref{fig:visual-sts} and Figure~\ref{fig:VDR_illustration}.

To further stress-test visual text understanding beyond English, we evaluate \ours on the cross-lingual and multilingual subsets of Visual STS. Cross-lingual results are summarized in Table~\ref{tab:sts_crosslingual_benchmarks} (full results in Table~\ref{appendix:tab:sts_crosslingual_benchmarks}), while multilingual results are reported in Table~\ref{appendix:tab:sts_multilingual_benchmarks} (Appendix~\ref{appendix:sec:comprehensive}).

\paragraph{Visual Semantic Textual Similarity (English)}

As shown in Table~\ref{tab:sts_benchmarks}, the variant of \ours pretrained only on the \textit{Mid-Training} datasets already achieves state-of-the-art performance across all English Visual STS tasks. Notably, it outperforms the largest existing vision encoders despite being $\sim$1/7 in model parameters and trained on approximately $\sim$1/87 examples seen.

Applying standard AllNLI fine-tuning with $\sim$270K examples further improves performance, yielding an additional $\sim$5-point gain in Spearman correlation. The strong performance of \ours on Visual STS \textbf{demonstrates its ability to capture semantic textual similarity directly from pixel inputs, highlighting effective zero-shot semantic understanding of rendered text}.

\paragraph{Visual Document Retrieval (VDR)}

Unlike prior VDR-focused models, \ours is primarily pretrained on synthetic visual text and does not explicitly train on real-world PDFs with complex document layouts. As such, its generalization to document retrieval tasks provides a stringent test of its visual text understanding capability.

Table~\ref{tab:VDR} reports performance of ViDoRe subsets~\cite{faysse2025colpali} in MIEB-lite \cite{xiao2025mieb}. Overall, \ours achieves state-of-the-art performance. In particular, it \textbf{exhibits strong capability in understanding \emph{tables and charts}, and documents where \emph{dense text is interleaved with structured visual elements}}, resulting in gains of $\sim$5-12 nDCG@5 on \texttt{AI} and \texttt{TabFQuAD}, and a substantial improvement of 16.6 nDCG@5 on \texttt{ShiftProject} over previous SOTA.

Importantly, \ours attains these results using a \textbf{vision-encoder-only} setup, in which textual queries are rendered as images and processed uniformly with visual documents. This setting places \ours at an inherent disadvantage relative to CLIP-style dual-encoder models, which benefit from a dedicated text encoder that provides semantically rich textual embeddings.

To quantify this gap, we conduct a fair comparison against \texttt{siglip-so400m-patch14-384}, the strongest SigLIP variant on ViDoRe. Shown in the middle section of Table~\ref{tab:VDR}, enforcing a unified visual processing pipeline for SigLIP leads to a substantial drop of 24.2 in nDCG@5 on average, underscoring the difficulty of performing VDR in a vision-only formulation and the robustness of \ours in visual text understanding.

\paragraph{Visual STS (Cross-lingual and Multilingual)}

\begin{figure*}[t]
    \centering
    \begin{subfigure}[b]{0.27\textwidth}
        \centering
        \includegraphics[width=\linewidth]{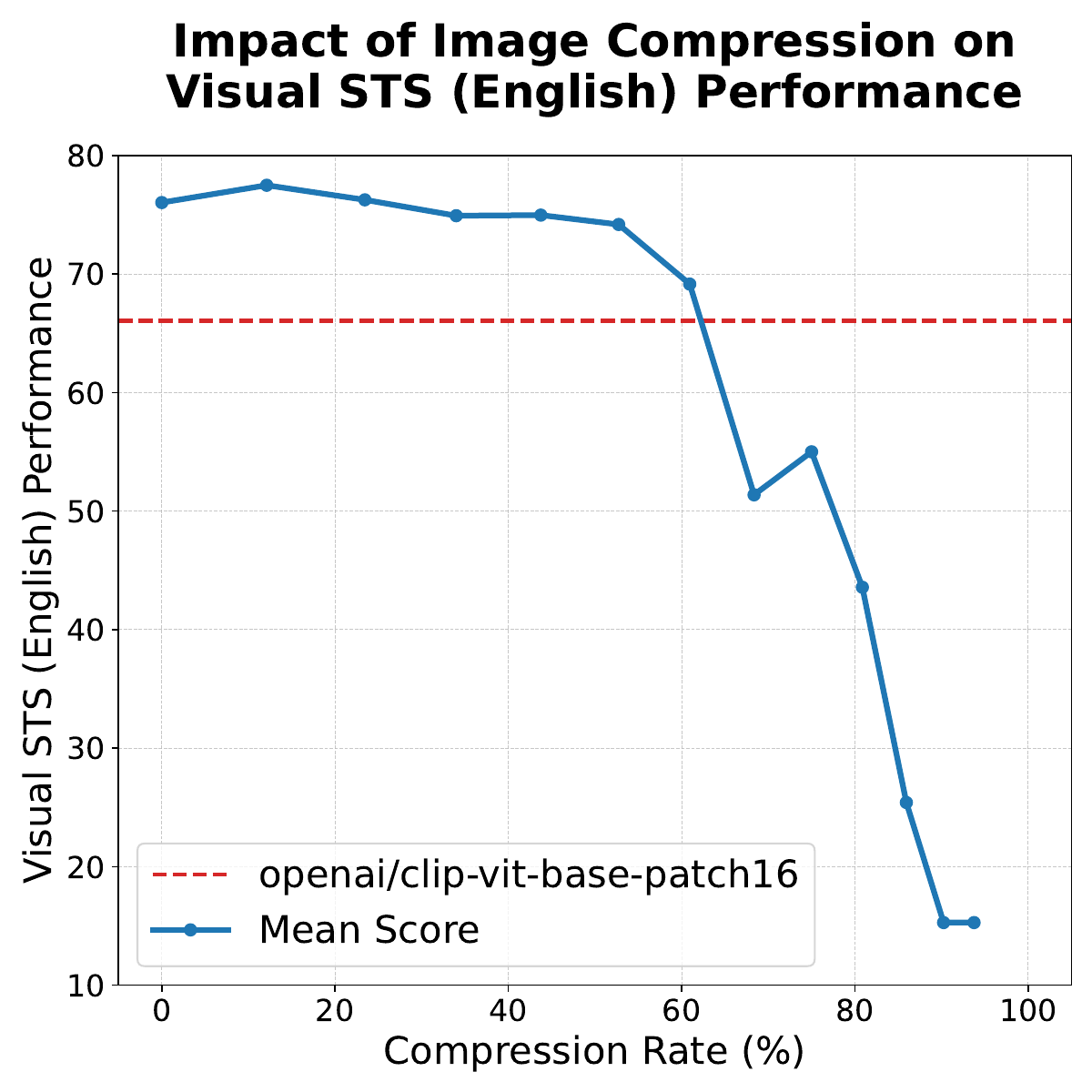}
        \label{fig:sts-english}
    \end{subfigure}
    \hfill
    \begin{subfigure}[b]{0.27\textwidth}
        \centering
        \includegraphics[width=\linewidth]{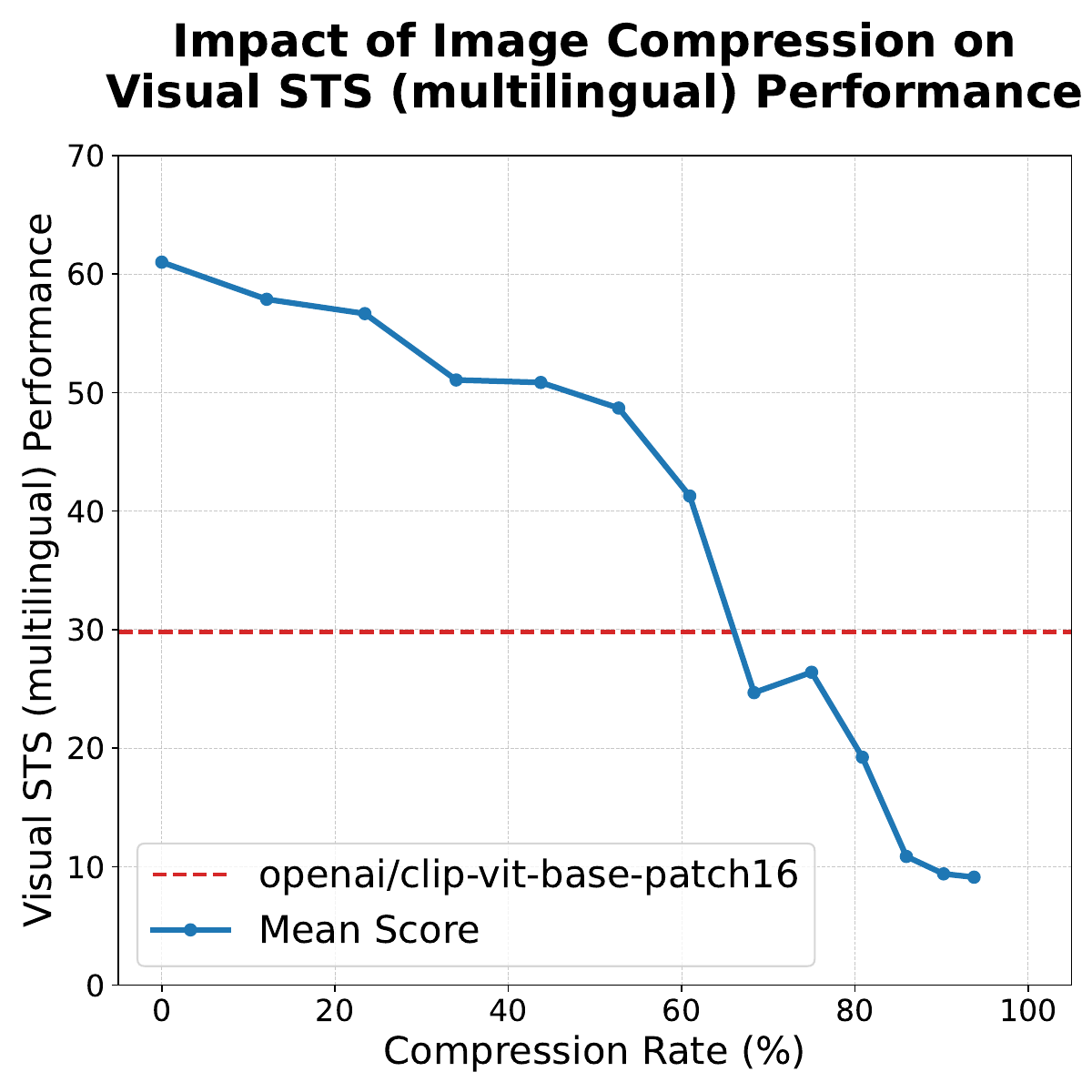}
        \label{fig:sts-multi}
    \end{subfigure}
    \hfill
    \begin{subfigure}[b]{0.27\textwidth}
        \centering
        \includegraphics[width=\linewidth]{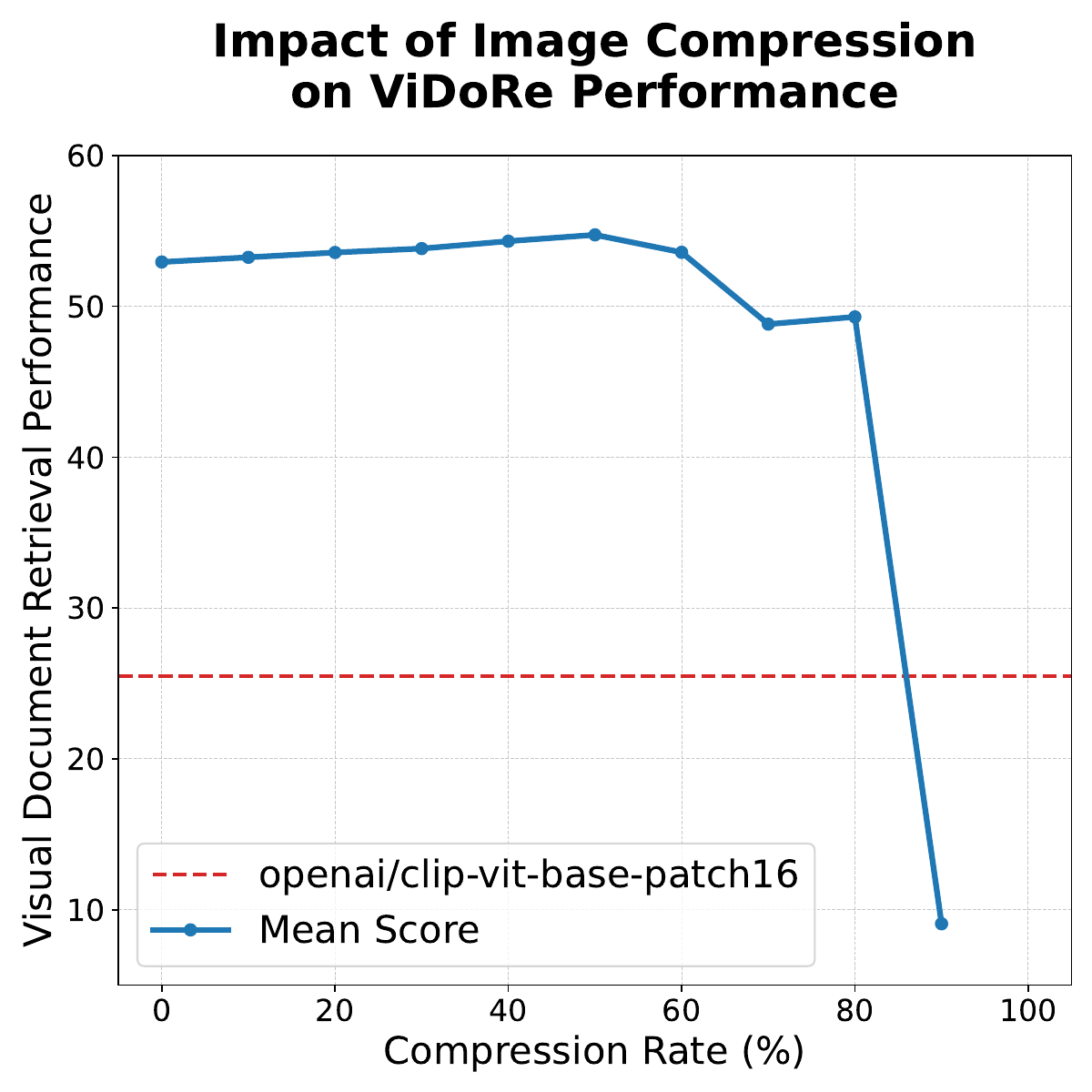}
        \label{fig:vidore}
    \end{subfigure}
    \caption{\ours performance under visual token compression.}
    \label{fig:compression}
\end{figure*}

\begin{figure*}[t]
    \centering
    \begin{subfigure}[b]{0.26\textwidth}
        \centering
        \includegraphics[width=\linewidth]{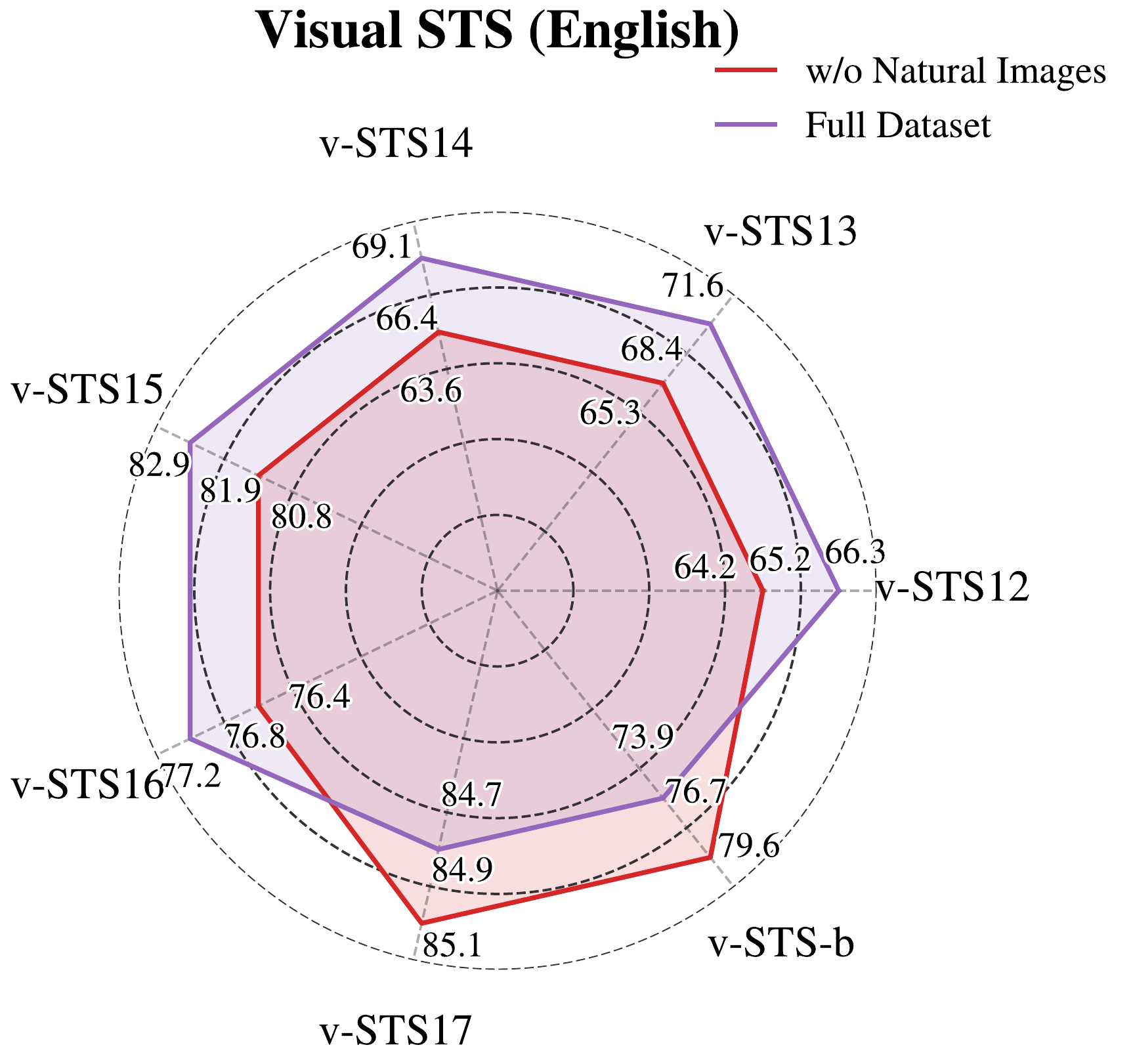}
        \label{fig:sub1}
    \end{subfigure}
    \hfill 
    \begin{subfigure}[b]{0.265\textwidth}
        \centering
        \includegraphics[width=\linewidth]{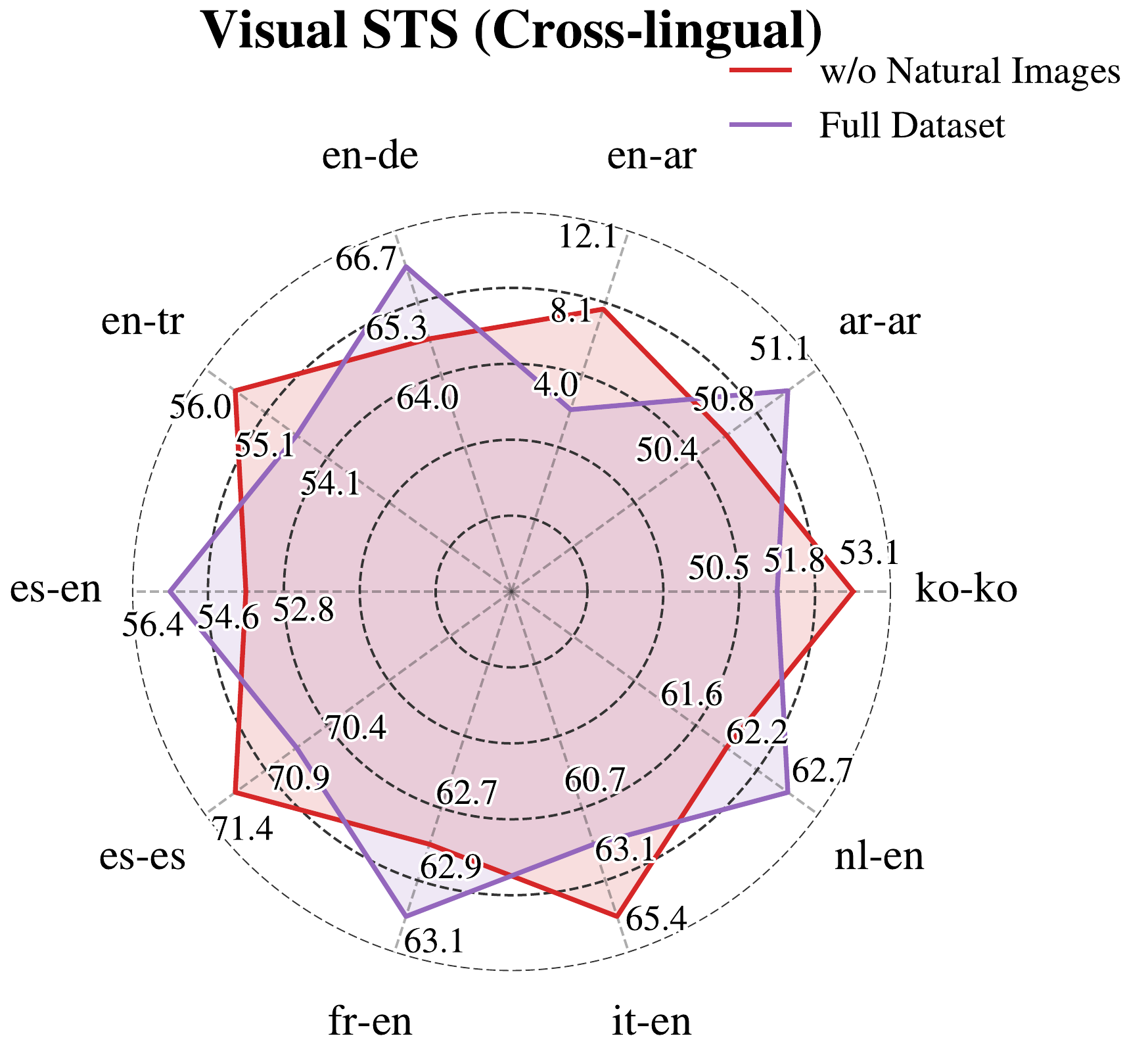}
        \label{fig:sub2}
    \end{subfigure}
    \hfill 
    \begin{subfigure}[b]{0.27\textwidth}
        \centering
        \includegraphics[width=\linewidth]{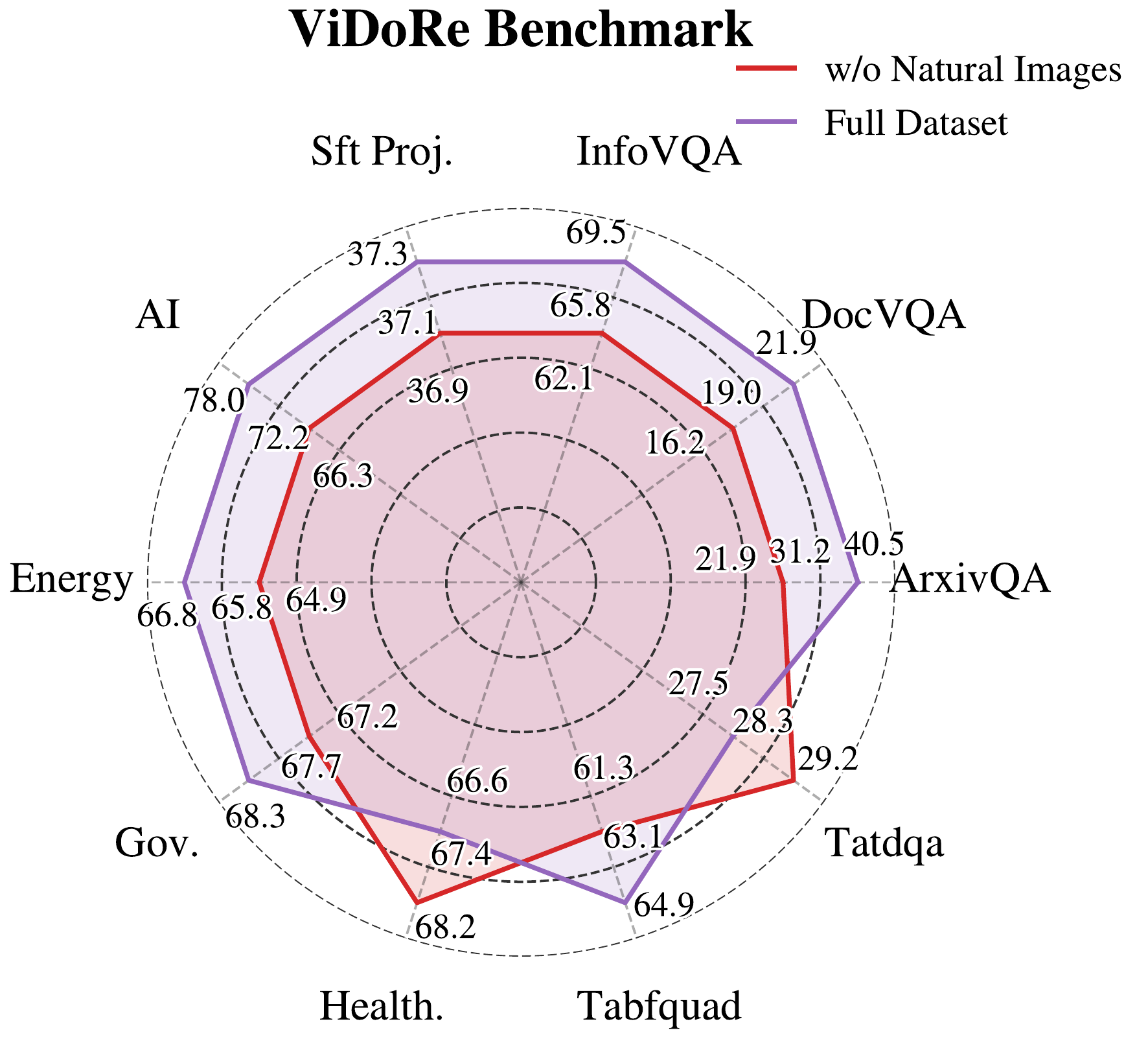}
        \label{fig:sub3}
    \end{subfigure}
    \caption{Performance comparison of models trained with and without natural images.}
    \label{fig:ablation_study_with_without_nautral_image}
\end{figure*}

Beyond English, \ours demonstrates strong and consistent capability in understanding multilingual text rendered as images. We evaluate this using the cross-lingual and multilingual Visual STS subsets derived from STS17 and STS-B, covering 11 languages. These include high-resource languages such as German, French, and Italian, as well as languages where prior vision encoders typically achieve near chance-level performance, such as Chinese, Russian, Korean, and Turkish.

Cross-lingual results comparing representative models are reported in Table~\ref{tab:sts_crosslingual_benchmarks} (More models in Table~\ref{appendix:tab:sts_crosslingual_benchmarks}) and multilingual results in Table~\ref{appendix:tab:sts_multilingual_benchmarks}. Even when trained solely on the \textit{Mid-Training} dataset, \ours achieves SOTA performance, \textbf{outperforming the strongest SigLIP variant by $\sim$15\% in Spearman correlation on cross-lingual tasks (Table~\ref{tab:sts_crosslingual_benchmarks}) and over 16\% on multilingual tasks (Table~\ref{appendix:tab:sts_multilingual_benchmarks})}.

Moreover, combining \textbf{pretraining + mid-training} yields consistently better performance than mid-training alone. 
This indicates \textbf{the importance of foundational multilingual knowledge through pretraining on massive unsupervised rendered corpus,
which is then effectively activated and enhanced through mid-training}, bringing multilingual visual text understanding closer to parity with English performance.

\paragraph{Evaluation on Downstream Tasks}

Described in Section 3.2, we evaluate \ours on MLLM downstream tasks.
Specifically, we compare with Qwen2.5-ViT~\cite{bai2025qwen2} initialized from Qwen2.5-VL-7B-Instruct, by paring both with the same LLM (Qwen2.5-7B-Instruct) and conducting LLaVA-style post-training. Our model achieves a 2.75\% average relative improvement over Qwen2.5-ViT across downstream tasks, 
validating its competence as a generalist vision encoder that provides comprehensive information to LLMs.
Detailed results on these benchmarks are in Table~\ref{tab:performance_comparison} in Appendix~\ref{appendix.mllm_evaluation}.

\section{In-depth Analysis}

\paragraph{Optical Context Compression}

Recent MLLMs utilize the visual modality as an efficient compression medium to alleviate textual context-length constraints \cite{wei2025deepseek, cheng2025glyph}. However, aggressive visual token downsampling risks severe semantic loss. To evaluate the semantic density of \ours, we downsample input images to induce varying compression rates across 32 Visual STS and ViDoRe tasks. We also evaluate visual token compression for MLLM tasks, further showing the strong potential of \ours to be integrated in modern MLLMs.

As shown in Figure~\ref{fig:compression}, our model retains strong semantic representation even under substantial compression. On Visual STS, \ours maintains performance parity with CLIP even when 60\% of the visual tokens are discarded (retaining only 118 of the original 196 tokens). This efficiency is even more pronounced on dense documents: on the ViDoRe benchmark, \ours continues to outperform the uncompressed CLIP baseline even at 80\% token compression. 

In Appendix~\ref{appendix.compression_mllm}, we further conduct a visual token compression sweep for MLLM downstream tasks, compared with the full-budget Qwen2.5-ViT baseline. As shown in Table~\ref{tab:mllm_compression}, with only 40\% visual tokens, \ours still outperforms the full-budget Qwen2.5-ViT in average.
These results confirm that our training recipe inherently yields highly compact representations, enabling extreme optical context compression without sacrificing downstream fidelity.

\paragraph{Validation of Multimodal Grounding at Scale}
\label{sec:ablation}

Having established in Section~\ref{sec: multimodal grounding} that purely synthetic pretraining triggers representation collapse, we investigate whether massive data scaling can overcome this limitation. We compare two full-scale variants of \ours pretrained on \textit{Text Corpus 2} with or without natural images (26M LAION pairs).

While this text-only variant matches the full model on fixed-resolution synthetic tasks like Visual STS (Figure~\ref{fig:ablation_study_with_without_nautral_image}), its performance drops substantially on the complex ViDoRe benchmark (Figure \ref{fig:ablation_study_with_without_nautral_image}). This full-scale degradation confirms that multimodal grounding cannot be bypassed through scale alone. Natural images act as a foundational regularizer; their diverse layouts, variable aspect ratios, and real-world contexts remain strictly required for robust document understanding at any scale.

\section{Related Work}

Traditional dual-encoder models like CLIP~\cite{radford2021learning} and SigLIP~\cite{zhai2023sigmoid,tschannen2025siglip} align images with tokenizer-based text encoders. Previous work demonstrated that language supervision injected certain OCR-related capability into CLIP vision encoders~\cite{tong2024cambrian,xiao2025mieb}. More recently, MLLM-based embedding models such as LCO-Embedding show that capabilities acquired during generative pretraining can be effectively activated through contrastive learning \cite{xiao2026scaling}. In parallel, vision-only approaches like PIXEL~\cite{rust2022language} and CLIPPO~\cite{tschannen2023clippo} model text visually. However, PIXEL relies on reconstruction objectives which lag in semantic discriminability, while CLIPPO lacks native resolution support essential for document processing.

Recent retrieval work has explored both architecture-side advances, such as 
vision-centric late-interaction retrievers like ColPali~\cite{faysse2025colpali}, 
and query-side adaptation to different retrieval environments~\cite{yuan2026understanding}. Web-SSL~\cite{fan2025scaling} scales unsupervised visual learning, proving on-par with language supervision. Distinct from these, \ours learns a unified, compact dense vector representation. It achieves high performance on representation benchmarks like ViDoRe while being able to serve as a generalist vision encoder to train MLLMs. Last but not least, \ours proves to serve as a robust and effective vision encoder for emerging trends of context compression~\cite{wei2025deepseek,cheng2025glyph}.

\section{Conclusion}

We presented \ours, a unified vision encoder for learning text representations directly from pixels. Rather than treating pixel-text modeling as a matter of scale alone, we identified four design fundamentals: spatial proxies from variable image resolutions and rendered font sizes, multimodal grounding with natural image-text pairs, layout-aware rendering to suppress visual shortcuts, and a multilingual curriculum that separates optical pretraining from semantic alignment.
\ours achieves state-of-the-art results on Visual STS and ViDoRe, transfers to downstream MLLM evaluation, and remains robust under aggressive visual token compression. 

\section*{Limitations}
Despite its strong performance, \ours is trained at a smaller data scale than many CLIP- and SigLIP-style baselines that rely on billion-scale image-text corpora such as LAION and DataComp~\cite{schuhmann2022laion,gadre2023datacomp}. Scaling the unified pixel-text training recipe, especially its natural image-text component, may further improve visual grounding. In addition, \ours is strongest on dense text and structured documents, but remains less competitive on diagram-heavy scientific subsets such as \texttt{ArxivQA}; this suggests that more diverse scientific figures, plots, and diagram-caption pairs would be useful pretraining data. Finally, our rendered text pairs provide controllable layout diversity, but may not cover all noise patterns in real-world text images, such as scans, blur, occlusion, handwriting, and low-quality camera captures.

\section*{Acknowledgment}
We would like to thank the anonymous reviewers and meta-reviewer for their valuable feedback on this work. This work was supported in part by the Multi-year Research Grant from the University of Macau (Grant No. MYRG-SRG2026-00032-FIC), and Research Grants Council of the Hong Kong SAR, China (No. CUHK 14206625).

\bibliography{custom}

@inproceedings{cimpoi2014describing,
  title={Describing textures in the wild},
  author={Cimpoi, Mircea and Maji, Subhransu and Kokkinos, Iasonas and Mohamed, Sammy and Vedaldi, Andrea},
  booktitle={Proceedings of the IEEE conference on computer vision and pattern recognition},
  pages={3606--3613},
  year={2014}
}

@article{bai2025qwen2,
  title={Qwen2.5-VL Technical Report}, 
  author={Shuai Bai and Keqin Chen and Xuejing Liu and Jialin Wang and Wenbin Ge and Sibo Song and Kai Dang and Peng Wang and Shijie Wang and Jun Tang and Humen Zhong and Yuanzhi Zhu and Mingkun Yang and Zhaohai Li and Jianqiang Wan and Pengfei Wang and Wei Ding and Zheren Fu and Yiheng Xu and Jiabo Ye and Xi Zhang and Tianbao Xie and Zesen Cheng and Hang Zhang and Zhibo Yang and Haiyang Xu and Junyang Lin},
  journal={arXiv preprint arXiv:2502.13923},
  year={2025}
}

@article{oord2018representation,
  title={Representation learning with contrastive predictive coding},
  author={Oord, Aaron van den and Li, Yazhe and Vinyals, Oriol},
  journal={arXiv preprint arXiv:1807.03748},
  year={2018}
}

@article{schuhmann2022laion,
  title={Laion-5b: An open large-scale dataset for training next generation image-text models},
  author={Schuhmann, Christoph and Beaumont, Romain and Vencu, Richard and Gordon, Cade and Wightman, Ross and Cherti, Mehdi and Coombes, Theo and Katta, Aarush and Mullis, Clayton and Wortsman, Mitchell and others},
  journal={Advances in neural information processing systems},
  volume={35},
  pages={25278--25294},
  year={2022}
}

@inproceedings{radford2021learning,
  title={Learning transferable visual models from natural language supervision},
  author={Radford, Alec and Kim, Jong Wook and Hallacy, Chris and Ramesh, Aditya and Goh, Gabriel and Agarwal, Sandhini and Sastry, Girish and Askell, Amanda and Mishkin, Pamela and Clark, Jack and others},
  booktitle={International conference on machine learning},
  pages={8748--8763},
  year={2021},
  organization={PmLR}
}

@inproceedings{tschannen2023clippo,
  title={Clippo: Image-and-language understanding from pixels only},
  author={Tschannen, Michael and Mustafa, Basil and Houlsby, Neil},
  booktitle={Proceedings of the IEEE/CVF Conference on Computer Vision and Pattern Recognition},
  pages={11006--11017},
  year={2023}
}

@article{tong2024cambrian,
  title={Cambrian-1: A fully open, vision-centric exploration of multimodal llms},
  author={Tong, Shengbang and Brown, Ellis and Wu, Penghao and Woo, Sanghyun and Middepogu, Manoj and Akula, Sai C and Yang, Jihan and Yang, Shusheng and Iyer, Adithya and Pan, Xichen and others},
  journal={Advances in Neural Information Processing Systems},
  volume={37},
  pages={87310--87356},
  year={2024}
}

@inproceedings{xiao2025mieb,
  title={Mieb: Massive image embedding benchmark},
  author={Xiao, Chenghao and Chung, Isaac and Kerboua, Imene and Stirling, Jamie and Zhang, Xin and Kardos, M{\'a}rton and Solomatin, Roman and Al Moubayed, Noura and Enevoldsen, Kenneth and Muennighoff, Niklas},
  booktitle={Proceedings of the IEEE/CVF International Conference on Computer Vision},
  pages={22187--22198},
  year={2025}
}

@article{xiao2024pixel,
  title={Pixel sentence representation learning},
  author={Xiao, Chenghao and Huang, Zhuoxu and Chen, Danlu and Hudson, G Thomas and Li, Yizhi and Duan, Haoran and Lin, Chenghua and Fu, Jie and Han, Jungong and Moubayed, Noura Al},
  journal={arXiv preprint arXiv:2402.08183},
  year={2024}
}

@article{dehghani2023patch,
  title={Patch n’pack: Navit, a vision transformer for any aspect ratio and resolution},
  author={Dehghani, Mostafa and Mustafa, Basil and Djolonga, Josip and Heek, Jonathan and Minderer, Matthias and Caron, Mathilde and Steiner, Andreas and Puigcerver, Joan and Geirhos, Robert and Alabdulmohsin, Ibrahim M and others},
  journal={Advances in Neural Information Processing Systems},
  volume={36},
  pages={2252--2274},
  year={2023}
}

@article{wei2025deepseek,
  title={Deepseek-ocr: Contexts optical compression},
  author={Wei, Haoran and Sun, Yaofeng and Li, Yukun},
  journal={arXiv preprint arXiv:2510.18234},
  year={2025}
}

@article{cheng2025glyph,
  title={Glyph: Scaling context windows via visual-text compression},
  author={Cheng, Jiale and Liu, Yusen and Zhang, Xinyu and Fei, Yulin and Hong, Wenyi and Lyu, Ruiliang and Wang, Weihan and Su, Zhe and Gu, Xiaotao and Liu, Xiao and others},
  journal={arXiv preprint arXiv:2510.17800},
  year={2025}
}

@inproceedings{mathew2022infographicvqa,
  title={Infographicvqa},
  author={Mathew, Minesh and Bagal, Viraj and Tito, Rub{\`e}n and Karatzas, Dimosthenis and Valveny, Ernest and Jawahar, CV},
  booktitle={Proceedings of the IEEE/CVF Winter Conference on Applications of Computer Vision},
  pages={1697--1706},
  year={2022}
}

@inproceedings{mathew2021docvqa,
  title={Docvqa: A dataset for vqa on document images},
  author={Mathew, Minesh and Karatzas, Dimosthenis and Jawahar, CV},
  booktitle={Proceedings of the IEEE/CVF winter conference on applications of computer vision},
  pages={2200--2209},
  year={2021}
}

@inproceedings{masry2022chartqa,
  title={Chartqa: A benchmark for question answering about charts with visual and logical reasoning},
  author={Masry, Ahmed and Do, Xuan Long and Tan, Jia Qing and Joty, Shafiq and Hoque, Enamul},
  booktitle={Findings of the association for computational linguistics: ACL 2022},
  pages={2263--2279},
  year={2022}
}

@inproceedings{singh2019towards,
  title={Towards vqa models that can read},
  author={Singh, Amanpreet and Natarajan, Vivek and Shah, Meet and Jiang, Yu and Chen, Xinlei and Batra, Dhruv and Parikh, Devi and Rohrbach, Marcus},
  booktitle={Proceedings of the IEEE/CVF conference on computer vision and pattern recognition},
  pages={8317--8326},
  year={2019}
}

@inproceedings{kembhavi2016diagram,
  title={A diagram is worth a dozen images},
  author={Kembhavi, Aniruddha and Salvato, Mike and Kolve, Eric and Seo, Minjoon and Hajishirzi, Hannaneh and Farhadi, Ali},
  booktitle={European conference on computer vision},
  pages={235--251},
  year={2016},
  organization={Springer}
}

@inproceedings{liu2024mmbench,
  title={Mmbench: Is your multi-modal model an all-around player?},
  author={Liu, Yuan and Duan, Haodong and Zhang, Yuanhan and Li, Bo and Zhang, Songyang and Zhao, Wangbo and Yuan, Yike and Wang, Jiaqi and He, Conghui and Liu, Ziwei and others},
  booktitle={European conference on computer vision},
  pages={216--233},
  year={2024},
  organization={Springer}
}

@inproceedings{li2023evaluating,
  title={Evaluating Object Hallucination in Large Vision-Language Models},
  author={Li, Yifan and Du, Yifan and Zhou, Kun and Wang, Jinpeng and Zhao, Wayne Xin and Wen, Ji-Rong},
  booktitle={Proceedings of the 2023 Conference on Empirical Methods in Natural Language Processing},
  pages={292--305},
  year={2023}
}

@article{chen2024we,
  title={Are we on the right way for evaluating large vision-language models?},
  author={Chen, Lin and Li, Jinsong and Dong, Xiaoyi and Zhang, Pan and Zang, Yuhang and Chen, Zehui and Duan, Haodong and Wang, Jiaqi and Qiao, Yu and Lin, Dahua and others},
  journal={Advances in Neural Information Processing Systems},
  volume={37},
  pages={27056--27087},
  year={2024}
}

@article{rust2022language,
  title={Language modelling with pixels},
  author={Rust, Phillip and Lotz, Jonas F and Bugliarello, Emanuele and Salesky, Elizabeth and de Lhoneux, Miryam and Elliott, Desmond},
  journal={arXiv preprint arXiv:2207.06991},
  year={2022}
}

@inproceedings{faysse2025colpali,
  title={Colpali: Efficient document retrieval with vision language models},
  author={Faysse, Manuel and Sibille, Hugues and Wu, Tony and Omrani, Bilel and Viaud, Gautier and Hudelot, C{\'e}line and Colombo, Pierre},
  booktitle={International Conference on Learning Representations},
  volume={2025},
  pages={61424--61449},
  year={2025}
}

@article{fan2025scaling,
  title={Scaling language-free visual representation learning},
  author={Fan, David and Tong, Shengbang and Zhu, Jiachen and Sinha, Koustuv and Liu, Zhuang and Chen, Xinlei and Rabbat, Michael and Ballas, Nicolas and LeCun, Yann and Bar, Amir and others},
  journal={arXiv preprint arXiv:2504.01017},
  year={2025}
}

@article{tschannen2025siglip,
  title={Siglip 2: Multilingual vision-language encoders with improved semantic understanding, localization, and dense features},
  author={Tschannen, Michael and Gritsenko, Alexey and Wang, Xiao and Naeem, Muhammad Ferjad and Alabdulmohsin, Ibrahim and Parthasarathy, Nikhil and Evans, Talfan and Beyer, Lucas and Xia, Ye and Mustafa, Basil and others},
  journal={arXiv preprint arXiv:2502.14786},
  year={2025}
}

@misc{qwen2025qwen25technicalreport,
      title={Qwen2.5 Technical Report}, 
      author={Qwen and : and An Yang and Baosong Yang and Beichen Zhang and Binyuan Hui and Bo Zheng and Bowen Yu and Chengyuan Li and Dayiheng Liu and Fei Huang and Haoran Wei and Huan Lin and Jian Yang and Jianhong Tu and Jianwei Zhang and Jianxin Yang and Jiaxi Yang and Jingren Zhou and Junyang Lin and Kai Dang and Keming Lu and Keqin Bao and Kexin Yang and Le Yu and Mei Li and Mingfeng Xue and Pei Zhang and Qin Zhu and Rui Men and Runji Lin and Tianhao Li and Tianyi Tang and Tingyu Xia and Xingzhang Ren and Xuancheng Ren and Yang Fan and Yang Su and Yichang Zhang and Yu Wan and Yuqiong Liu and Zeyu Cui and Zhenru Zhang and Zihan Qiu},
      year={2025},
      eprint={2412.15115},
      archivePrefix={arXiv},
      primaryClass={cs.CL},
      url={https://arxiv.org/abs/2412.15115}, 
}

@inproceedings{zhang2025lmms,
  title={Lmms-eval: Reality check on the evaluation of large multimodal models},
  author={Zhang, Kaichen and Li, Bo and Zhang, Peiyuan and Pu, Fanyi and Cahyono, Joshua Adrian and Hu, Kairui and Liu, Shuai and Zhang, Yuanhan and Yang, Jingkang and Li, Chunyuan and others},
  booktitle={Findings of the Association for Computational Linguistics: NAACL 2025},
  pages={881--916},
  year={2025}
}

@article{ilharco2021openclip,
  title={Openclip},
  author={Ilharco, Gabriel and Wortsman, Mitchell and Carlini, Nicholas and Taori, Rohan and Dave, Achal and Shankar, Vaishaal and Namkoong, Hongseok and Miller, John and Hajishirzi, Hannaneh and Farhadi, Ali and others},
  journal={Zenodo},
  year={2021}
}

@article{gadre2023datacomp,
  title={Datacomp: In search of the next generation of multimodal datasets},
  author={Gadre, Samir Yitzhak and Ilharco, Gabriel and Fang, Alex and Hayase, Jonathan and Smyrnis, Georgios and Nguyen, Thao and Marten, Ryan and Wortsman, Mitchell and Ghosh, Dhruba and Zhang, Jieyu and others},
  journal={Advances in Neural Information Processing Systems},
  volume={36},
  pages={27092--27112},
  year={2023}
}

@inproceedings{zhai2023sigmoid,
  title={Sigmoid loss for language image pre-training},
  author={Zhai, Xiaohua and Mustafa, Basil and Kolesnikov, Alexander and Beyer, Lucas},
  booktitle={Proceedings of the IEEE/CVF international conference on computer vision},
  pages={11975--11986},
  year={2023}
}

@article{sun2023eva,
  title={Eva-clip: Improved training techniques for clip at scale},
  author={Sun, Quan and Fang, Yuxin and Wu, Ledell and Wang, Xinlong and Cao, Yue},
  journal={arXiv preprint arXiv:2303.15389},
  year={2023}
}

@article{izacard2021unsupervised,
  title={Unsupervised dense information retrieval with contrastive learning},
  author={Izacard, Gautier and Caron, Mathilde and Hosseini, Lucas and Riedel, Sebastian and Bojanowski, Piotr and Joulin, Armand and Grave, Edouard},
  journal={arXiv preprint arXiv:2112.09118},
  year={2021}
}

@misc{grokv2024,
  title={Grok-1.5 Vision Preview},
  author={xAI},
  year={2024},
  url={https://x.ai/blog/grok-1.5v}
}

@inproceedings{yuan2026understanding,
  title={Understanding the Behaviors of Environment-aware Information Retrieval},
  author={Yuan, Ruifeng and Yuan, Chaohao and Dai, David and Rong, Yu and Cheng, Hong and Chan, Hou Pong and Xiao, Chenghao},
  booktitle={Proceedings of the 64th Annual Meeting of the Association for Computational Linguistics (Volume 1: Long Papers)},
  pages={43490--43503},
  year={2026}
}

@article{xiao2026scaling,
  title={Scaling language-centric omnimodal representation learning},
  author={Xiao, Chenghao and Chan, Hou Pong Ken and Zhang, Hao and Xu, Weiwen and Aljunied, Mahani and Rong, Yu},
  journal={Advances in Neural Information Processing Systems},
  volume={38},
  pages={158370--158401},
  year={2026}
}

@inproceedings{shabtay2025livexiv,
  title={Livexiv-a multi-modal live benchmark based on arxiv papers content},
  author={Shabtay, Nimrod and Maia Polo, Felipe and Doveh, Sivan and Lin, Wei and Mirza, Muhammad Jehanzeb and Choshen, Leshem and Yurochkin, Mikhail and Sun, Yuekai and Arbelle, Assaf and Karlinsky, Leonid and others},
  booktitle={International Conference on Learning Representations},
  volume={2025},
  pages={11470--11502},
  year={2025}
}

\newpage
\appendix

\section{Fonts}
\label{appendix.fonts}

We utilize a diverse set of fonts to ensure the robustness of our rendering pipeline. The font library consists of 393 unique font files spanning multiple scripts and weights (100-900). We distribute more kinds of fonts to the language that occupies a larger portion in the dataset. A detailed breakdown of the font families is provided in Table~\ref{tab:font_lib}.

\begin{table}[h]
\centering
\resizebox{0.5\textwidth}{!}{%
\begin{tabular}{@{}lp{10cm}@{}}
\toprule
\textbf{Category} & \textbf{Font Families} \\ \midrule
\textbf{Chinese} & Noto Sans/Serif SC, Liu Jian Mao Cao, Long Cang, Ma Shan Zheng, Zhi Mang Xing, ZCOOL Series \\
\textbf{Japanese} & Noto Sans/Serif JP, DotGothic16, Kiwi Maru, Potta One, Reggae One, RocknRoll One \\
\textbf{Korean} & Noto Sans/Serif KR, Gothic A1, Do Hyeon, Jua, Yeon Sung, Nanum Series \\
\textbf{Arabic} & Noto Sans Arabic, Amiri, Cairo \\
\textbf{English} & Roboto, Open Sans, Lato, Montserrat, Nunito, Playfair Display, Poppins, Quicksand, Raleway, PT Sans, Ubuntu, Lora, Merriweather, Libre Baskerville, Anton, Josefin Sans, Caveat, Dancing Script, Pacifico, Shadows Into Light, Great Vibes, Allura, Cookie, Courgette, Lobster, Parisienne, Sacramento, Satisfy, Tangerine, Yellowtail \\
\textbf{Code} & JetBrains Mono, Fira Code, Roboto Mono, Source Code Pro, Ubuntu Mono, Inconsolata, Space Mono \\
\textbf{Others} & Noto Sans variants covering: Armenian, Bengali, Devanagari, Ethiopic, Georgian, Gujarati, Gurmukhi, Hebrew, Kannada, Khmer, Lao, Malayalam, Math, Mongolian, Myanmar, Tamil, Telugu, Thai, and Symbols. \\ \bottomrule
\end{tabular}%
}
\caption{Summary of Font Families}
\label{tab:font_lib}
\end{table}

\section{Training Configuration}
\label{appendix.training_config}

\subsection{Render Engine in Pretraining}

Table~\ref{tab:aug_params} summarizes the quantitative configurations used in our data generation engine.

\begin{table}[h]
    \centering
    \resizebox{0.5\textwidth}{!}
    {\begin{tabular}{l|c|l}
        \toprule
        \textbf{Parameter} & \textbf{Value/Range} & \textbf{Description} \\
        \midrule
        Canvas Size & $224 \times 224$ & Fixed input resolution \\
        Font Size & $U(16, 28)$ & Sampled uniformly \\
        Max Lines & 12 & Text wrapping limit \\
        Background Prob. ($p_{\text{bg}}$) & 0.5 & Probability of using DTD textures \\
        Rotation Angle & $[-15^{\circ}, +15^{\circ}]$ & Random rotation \\
        Position Jitter & $\pm 20$ pixels & Random $(x, y)$ shift from center \\
        Blur Prob. ($p_{\text{blur}}$) & 0.2 & Probability of Gaussian Blur \\
        Blur Radius & $[0.5, 1.2]$ & Strength of blur \\
        Stroke Prob. ($p_{\text{stroke}}$) & 0.4 & Probability of adding text outline \\
        Brightness Jitter & $[0.6, 1.4]$ & Applied to background images \\
        \bottomrule
    \end{tabular}}
    \caption{Hyperparameters for On-the-fly Text Rendering.}
    \label{tab:aug_params}
\end{table}

\begin{table*}[t]
    \centering
    \resizebox{\textwidth}{!}{%
        \begin{tabular}{lccccccccc}
            \toprule
            \textbf{Model} & \textbf{InfoVQA} & \textbf{DocVQA} & \textbf{TextVQA} & \textbf{LiveXIVVQA} & \textbf{AI2D} & \textbf{MMB\textsubscript{en}} & \textbf{POPE} & \textbf{RealWorldQA} & \textbf{MMStar} \\
            \midrule
            \ours   & \textbf{31.1} & \textbf{72.0} & \textbf{63.3} & \textbf{44.6} & \textbf{77.3} & \textbf{67.2} & \textbf{86.9} & \textbf{60.0} & \textbf{46.7} \\
            Qwen2.5-ViT & 28.7          & 71.3          & 63.1          & 44.4          & 75.9          & 64.9          & 86.6          & 56.6          & 45.4          \\
            \bottomrule
        \end{tabular}%
    }
    \caption{Performance comparison on downstream tasks under MLLM evaluation.}
    \label{tab:performance_comparison}
\end{table*}

\subsection{Training Setting in End-to-end Evaluation on Downstream Tasks}
\label{appendix.mllm_evaluation}

In our experiments, we utilize various ViT architectures as visual encoders, paired with Qwen2.5-7b-Instruct~\cite{qwen2025qwen25technicalreport} as the LLM backbone. Following the training paradigm proposed in LLaVA, the training process is divided into two stages:

Stage 1: Only the projector is trainable (the backbones are frozen). We train the model for one epoch with a learning rate of $2.0 \times 10^{-4}$ and a batch size of 128. A cosine learning rate schedule with a warm-up ratio of 0.1 is applied.

Stage 2: The entire model is fully fine-tuned. We train for three epochs with a learning rate of $2.0 \times 10^{-5}$ and a batch size of 8, using 4 gradient accumulation steps. Consistent with the first stage, we employ a cosine scheduler with a warm-up ratio of 0.1.

We evaluate the resulting MLLMs on a suite of widely adopted benchmarks covering both text-centric and general multimodal understanding using the LMMs-Eval framework~\cite{zhang2025lmms}. Specifically, \textbf{OCR and document understanding tasks} include InfoVQA \cite{mathew2022infographicvqa}, DocVQA \cite{mathew2021docvqa}, ChartQA \cite{masry2022chartqa}, TextVQA \cite{singh2019towards}, and LiveXivVQA \cite{shabtay2025livexiv}. \textbf{General vision understanding tasks} include AI2D \cite{kembhavi2016diagram}, MMBench\textsuperscript{EN} \cite{liu2024mmbench}, POPE \cite{li2023evaluating}, RealWorldQA \cite{grokv2024}, and MMStar \cite{chen2024we}.

Results are summarized in Table~\ref{tab:performance_comparison}, where \ours outperforms Qwen2.5-ViT when used as the vision encoder in end-to-end MLLM training and evaluation, across 9 tasks.

\section{Compression on MLLM tasks}
\label{appendix.compression_mllm}
Table~\ref{tab:mllm_compression} shows the results of a full 10\%-90\% visual token compression sweep, with the full-budget Qwen2.5-ViT scores shown in the leftmost column for direct per-task comparison. The overall trend remains favorable under compression: \ours reaches a 60.04 mean score at 40\% token keep (2.5x compression), 60.61 at 50\% keep (2.0x), and 60.93 at 60\% keep (1.67x), compared to 59.67 for full-budget Qwen2.5-ViT. At 50\% keep, all 9 tasks are within 95\% of the corresponding Qwen score, and 6 out of 9 tasks surpass the full-budget baseline outright; this rises to 7 out of 9 tasks at 60\% keep. Recovery varies slightly by task: most datasets reach or exceed the Qwen reference by 30\%–50\% keep. TextVQA remains the most compression-sensitive, which is expected as its examples typically contain very small text embedded in natural images, making them inherently challenging to compress. Overall, these results indicate that \ours preserves its downstream advantage even under substantial visual token reduction, rather than benefiting only from a larger visual token budget.

\begin{table*}[t]
\centering 
\footnotesize
\setlength{\tabcolsep}{3.5pt}
\renewcommand{\arraystretch}{1.15}
\begin{tabular}{lc|ccccccccc}
\toprule
 & & \multicolumn{9}{c}{\ours under visual token compression} \\
\cmidrule(lr){3-11}
\textbf{Dataset} & \makecell{Qwen2.5-ViT\\ }
 & \makecell{10\%\\ ($10.0\times$)} & \makecell{20\%\\ ($5.0\times$)}
 & \makecell{30\%\\ ($3.33\times$)} & \makecell{40\%\\ ($2.5\times$)}
 & \makecell{50\%\\ ($2.0\times$)} & \makecell{60\%\\ ($1.67\times$)}
 & \makecell{70\%\\ ($1.43\times$)} & \makecell{80\%\\ ($1.25\times$)}
 & \makecell{90\%\\ ($1.11\times$)} \\
\midrule
AI2D         & 75.91 & 71.31 & 74.51 & \textbf{76.10} & 75.71 & 75.58 & \textbf{76.75} & \textbf{76.42} & \textbf{76.36} & \textbf{76.68} \\
DocVQA       & 71.32 & 67.83 & \textbf{74.48} & \textbf{75.41} & \textbf{75.01} & \textbf{75.19} & \textbf{74.83} & \textbf{74.06} & \textbf{73.42} & \textbf{72.57} \\
InfoVQA      & 28.72 & 22.13 & 26.01 & \textbf{28.99} & \textbf{30.00} & \textbf{30.86} & \textbf{30.90} & \textbf{31.55} & \textbf{31.11} & \textbf{31.30} \\
LiveXiv-VQA  & 44.42 & 38.85 & 41.87 & 43.98 & \textbf{44.47} & \textbf{45.02} & \textbf{44.88} & \textbf{45.13} & \textbf{44.68} & \textbf{44.68} \\
MMBench-EN   & 64.95 & 61.17 & 64.69 & \textbf{66.32} & \textbf{66.92} & \textbf{66.32} & \textbf{66.58} & \textbf{67.18} & \textbf{66.92} & \textbf{67.87} \\
MMStar       & 45.40 & 41.36 & 42.77 & 44.43 & 45.27 & \textbf{46.56} & \textbf{46.59} & \textbf{47.08} & 45.16 & \textbf{46.90} \\
POPE         & 86.64 & 78.46 & 79.26 & 83.52 & 84.41 & 84.78 & 86.07 & 86.54 & 86.50 & \textbf{87.07} \\
RealWorldQA  & 56.60 & 55.82 & \textbf{57.25} & \textbf{60.39} & \textbf{60.78} & \textbf{60.92} & \textbf{61.05} & \textbf{60.78} & \textbf{60.39} & 
\textbf{61.44} \\
TextVQA      & 63.11 & 39.07 & 48.67 & 55.14 & 57.73 & 60.22 & 60.70 & 61.90 & 62.53 & 62.86 \\
\midrule
\textbf{Avg.} & 59.67 & 52.89 & 56.61 & 59.37 & \textbf{60.04} & \textbf{60.61} & \textbf{60.93} & \textbf{61.18} & \textbf{60.79} & \textbf{61.26} \\
$\Delta$ vs.\ Qwen & -- & $-6.79$ & $-3.06$ & $-0.31$ & $+0.36$ & $+0.93$ & $+1.26$ & $+1.51$ & $+1.11$ & $+1.59$ \\
\bottomrule
\end{tabular}
\caption{Downstream MLLM performance of \ours under visual token compression. Each column reports a visual-token keep ratio with the corresponding compression factor in parentheses; the second column is the uncompressed Qwen2.5-ViT. \textbf{Bold} denotes settings where \ours with compressed contexts outperforms full-budget Qwen2.5-ViT.}
\label{tab:mllm_compression}
\end{table*}

\section{Examples of Rendered Images}
\label{appendix.examples}

\begin{figure*}[p]
    \centering

    \begin{subfigure}[b]{0.22\textwidth}
        \includegraphics[width=\linewidth, height=\linewidth]{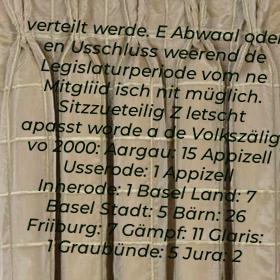}
    \end{subfigure}
    \begin{subfigure}[b]{0.22\textwidth}
        \includegraphics[width=\linewidth, height=\linewidth]{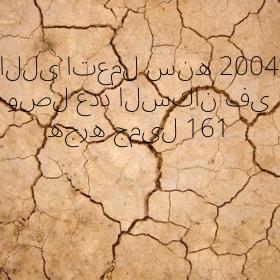}
    \end{subfigure}
    \begin{subfigure}[b]{0.22\textwidth}
        \includegraphics[width=\linewidth, height=\linewidth]{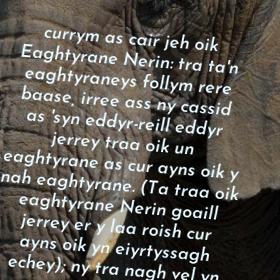}
    \end{subfigure}
    \begin{subfigure}[b]{0.22\textwidth}
        \includegraphics[width=\linewidth, height=\linewidth]{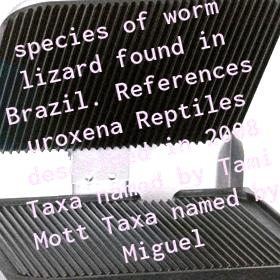}
    \end{subfigure}
    
    \begin{subfigure}[b]{0.22\textwidth}
        \includegraphics[width=\linewidth, height=\linewidth]{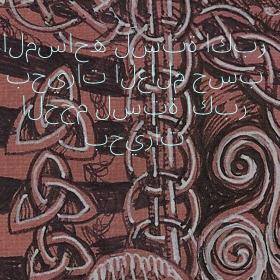}
    \end{subfigure}
    \begin{subfigure}[b]{0.22\textwidth}
        \includegraphics[width=\linewidth, height=\linewidth]{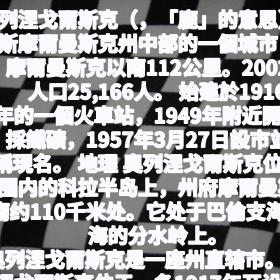}
    \end{subfigure}
    \begin{subfigure}[b]{0.22\textwidth}
        \includegraphics[width=\linewidth, height=\linewidth]{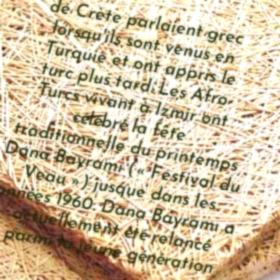}
    \end{subfigure}
    \begin{subfigure}[b]{0.22\textwidth}
        \includegraphics[width=\linewidth, height=\linewidth]{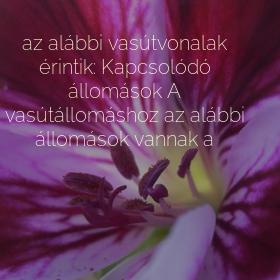}
    \end{subfigure}
    \begin{subfigure}[b]{0.22\textwidth}
        \includegraphics[width=\linewidth, height=\linewidth]{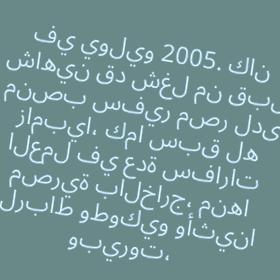}
    \end{subfigure}
    \begin{subfigure}[b]{0.22\textwidth}
        \includegraphics[width=\linewidth, height=\linewidth]{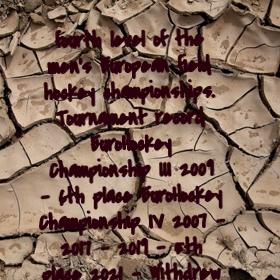}
    \end{subfigure}
    \begin{subfigure}[b]{0.22\textwidth}
        \includegraphics[width=\linewidth, height=\linewidth]{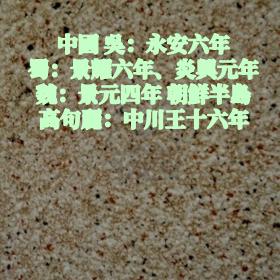}
    \end{subfigure}
    \begin{subfigure}[b]{0.22\textwidth}
        \includegraphics[width=\linewidth, height=\linewidth]{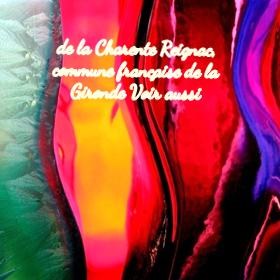}
    \end{subfigure}
    \begin{subfigure}[b]{0.22\textwidth}
        \includegraphics[width=\linewidth, height=\linewidth]{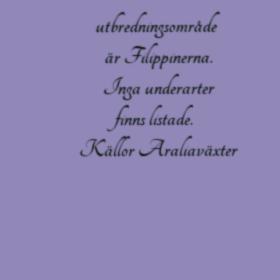}
    \end{subfigure}
    \begin{subfigure}[b]{0.22\textwidth}
        \includegraphics[width=\linewidth, height=\linewidth]{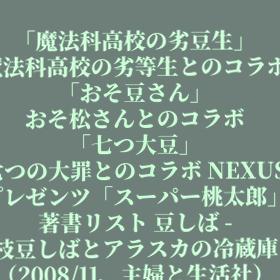}
    \end{subfigure}
    \begin{subfigure}[b]{0.22\textwidth}
        \includegraphics[width=\linewidth, height=\linewidth]{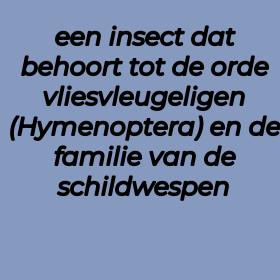}
    \end{subfigure}
    \begin{subfigure}[b]{0.22\textwidth}
        \includegraphics[width=\linewidth, height=\linewidth]{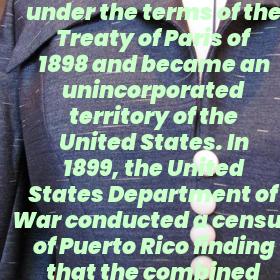}
    \end{subfigure}
    \begin{subfigure}[b]{0.22\textwidth}
        \includegraphics[width=\linewidth, height=\linewidth]{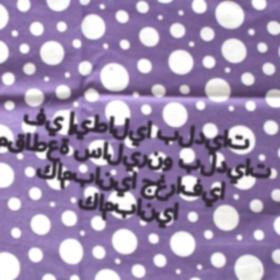}
    \end{subfigure}
    \begin{subfigure}[b]{0.22\textwidth}
        \includegraphics[width=\linewidth, height=\linewidth]{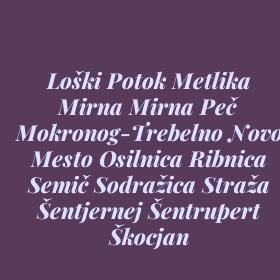}
    \end{subfigure}
    \begin{subfigure}[b]{0.22\textwidth}
        \includegraphics[width=\linewidth, height=\linewidth]{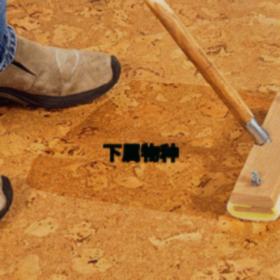}
    \end{subfigure}
    \begin{subfigure}[b]{0.22\textwidth}
        \includegraphics[width=\linewidth, height=\linewidth]{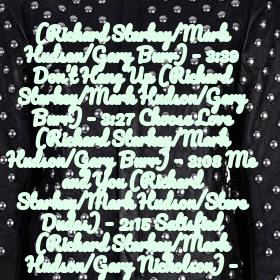}
    \end{subfigure}
    \begin{subfigure}[b]{0.22\textwidth}
        \includegraphics[width=\linewidth, height=\linewidth]{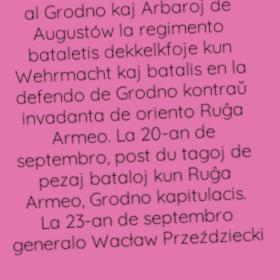}
    \end{subfigure}
    \begin{subfigure}[b]{0.22\textwidth}
        \includegraphics[width=\linewidth, height=\linewidth]{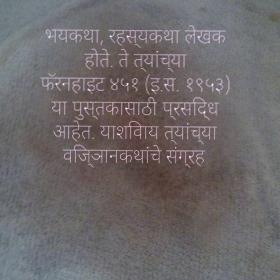}
    \end{subfigure}
    \begin{subfigure}[b]{0.22\textwidth}
        \includegraphics[width=\linewidth, height=\linewidth]{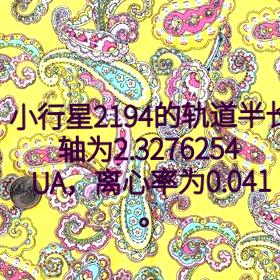}
    \end{subfigure}
    \begin{subfigure}[b]{0.22\textwidth}
        \includegraphics[width=\linewidth, height=\linewidth]{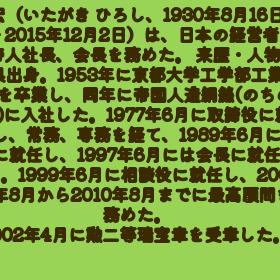}
    \end{subfigure}
    \caption{Examples of rendered images}
    \label{fig:examples}
\end{figure*}

We provide examples of multilingual rendered texts from the dataset in Figure~\ref{fig:examples}. 

\section{Comprehensive Comparison with Baselines}
\label{appendix:sec:comprehensive}

The comparison with full baselines are listed in Table~\ref{appendix:tab:sts_benchmarks}, Table~\ref{appendix:tab:VDR}, Table~\ref{appendix:tab:sts_crosslingual_benchmarks}, and Table~\ref{appendix:tab:sts_multilingual_benchmarks}.

\begin{table*}[ht]
\centering
\resizebox{\textwidth}{!}{%
\begin{tabular}{lcccccccc}
\toprule
\textbf{Model name} & \textbf{v-STS12} & \textbf{v-STS13} & \textbf{v-STS14} & \textbf{v-STS15} & \textbf{v-STS16} & \textbf{v-STS17} & \textbf{v-STS-b} & \textbf{Avg.} \\
\midrule
google/siglip-base-patch16-224 & 63.19 & 55.40 & 57.99 & 73.07 & 67.79 & 77.78 & 54.50 & 64.25 \\
openai/clip-vit-large-patch14 & 53.89 & 66.78 & 55.98 & 72.03 & 70.49 & 75.26 & 56.74 & 64.45 \\
google/siglip-base-patch16-256-multilingual & 66.62 & 54.80 & 59.00 & 72.65 & 68.33 & 80.53 & 56.29 & 65.46 \\
laion/CLIP-ViT-H-14-laion2B-s32B-b79K & 57.00 & 62.25 & 58.62 & 74.40 & 70.57 & 76.69 & 58.99 & 65.50 \\
laion/CLIP-ViT-L-14-laion2B-s32B-b82K & 57.52 & 62.75 & 59.94 & 74.55 & 70.61 & 75.92 & 59.43 & 65.82 \\
openai/clip-vit-base-patch16 & 63.82 & 63.26 & 56.99 & 73.32 & 68.91 & 78.18 & 57.93 & 66.06 \\
google/siglip-base-patch16-256 & 65.01 & 58.02 & 60.36 & 74.25 & 69.09 & 78.73 & 57.65 & 66.16 \\
google/siglip-base-patch16-384 & 64.62 & 59.38 & 61.17 & 74.34 & 70.29 & 79.27 & 60.28 & 67.05 \\
google/siglip-large-patch16-256 & 63.94 & 59.44 & 59.35 & 75.74 & 71.83 & 79.21 & 62.50 & 67.43 \\
google/siglip-base-patch16-512 & 64.97 & 59.10 & 61.13 & 75.08 & 71.27 & 80.09 & 62.21 & 67.69 \\
google/siglip-so400m-patch14-384 & 61.90 & 62.95 & 60.58 & 76.17 & 73.48 & 78.41 & 62.63 & 68.02 \\
laion/CLIP-ViT-B-16-DataComp.XL-s13B-b90K & 64.19 & 63.81 & 62.34 & 75.48 & 69.90 & 80.04 & 63.51 & 68.47 \\
EVA02-CLIP-bigE-14 & 62.24 & 62.36 & 62.17 & 77.41 & 73.63 & 80.96 & 62.85 & 68.80 \\
laion/CLIP-ViT-g-14-laion2B-s34B-b88K & 61.85 & 66.43 & 62.32 & 76.73 & 72.67 & 79.88 & 64.13 & 69.14 \\
google/siglip-large-patch16-384 & 66.30 & 62.08 & 61.66 & 77.11 & 73.27 & 79.58 & 66.59 & 69.51 \\
laion/CLIP-ViT-L-14-DataComp.XL-s13B-b90K & 62.36 & 67.64 & 64.25 & 77.36 & 73.48 & 80.63 & 63.38 & 69.87 \\
laion/CLIP-ViT-bigG-14-laion2B-39B-b160k & 62.81 & 68.16 & 65.50 & 78.67 & 74.89 & 79.97 & 66.54 & 70.93 \\
EVA02-CLIP-bigE-14-plus & 63.36 & 68.00 & 66.38 & 79.45 & 75.26 & 82.87 & 68.59 & 71.99 \\
\midrule
\multicolumn{9}{c}{\textbf{\textit{Backbone}}} \\
Qwen2.5-VIT & 47.50 & 36.49 & 30.95 & 54.69 & 53.71 & 63.87 & 38.63 & 46.55\\
\multicolumn{9}{c}{\textbf{\textit{Ours}}} \\
\ours (mid-training only) & 65.78 & 70.00 & 67.76 & 82.39 & 76.99 & 84.83 & 75.30 & 74.72\\ 
\ours (mid-training + finetuned) & 76.60 & 75.94 & 75.07 & 85.17 & 79.65 & 85.25 & 80.93 & 79.80\\
\bottomrule
\end{tabular}%
}
\caption{\ours Encoder Performance on Visual STS Tasks (English-only), which renders traditional STS tasks in NLP as image-only tasks, assessing vision models' text-on-image semantic understanding.}
\label{appendix:tab:sts_benchmarks}
\end{table*}

\begin{table*}[ht]
\centering
\resizebox{\textwidth}{!}{%
\begin{tabular}{lccccccccccc}
\toprule
\textbf{Model name} & \textbf{ArxivQA} & \textbf{DocVQA} & \textbf{InfoVQA} & \textbf{Sft Proj.} & \textbf{AI} & \textbf{Energy} & \textbf{Gov.} & \textbf{Health.} & \textbf{Tabfquad} & \textbf{Tatdqa} & \textbf{Avg.} \\
\midrule
\multicolumn{12}{c}{\textbf{\textit{Baselines}}} \\
openai/clip-vit-base-patch16 & 26.54 & 14.60 & 51.70 & 7.13 & 22.86 & 32.43 & 39.84 & 37.54 & 17.61 & 4.71 & 25.50 \\
google/siglip-base-patch16-224 & 31.49 & 16.04 & 46.11 & 3.71 & 25.27 & 35.53 & 32.35 & 37.01 & 29.04 & 5.08 & 26.16 \\
laion/CLIP-ViT-B-16-DataComp.XL-s13B-b90K & 28.88 & 13.97 & 46.88 & 7.25 & 32.17 & 38.53 & 31.05 & 35.83 & 26.60 & 9.07 & 27.02 \\
EVA02-CLIP-bigE-14 & 32.72 & 16.35 & 54.80 & 10.14 & 33.53 & 48.50 & 41.32 & 42.98 & 28.80 & 7.09 & 31.62 \\
google/siglip-base-patch16-256 & 35.17 & 19.42 & 48.73 & 5.45 & 31.06 & 41.28 & 40.07 & 49.94 & 37.00 & 8.50 & 31.66 \\
EVA02-CLIP-bigE-14-plus & 34.86 & 16.84 & 55.19 & 12.76 & 34.57 & 44.99 & 43.14 & 42.47 & 30.36 & 7.52 & 32.27 \\
openai/clip-vit-large-patch14 & 28.64 & 16.69 & 62.44 & 17.05 & 38.25 & 61.62 & 52.84 & 60.23 & 30.95 & 11.00 & 37.97 \\
laion/CLIP-ViT-L-14-DataComp.XL-s13B-b90K & 34.51 & 19.68 & 55.61 & 16.19 & 47.20 & 58.93 & 50.28 & 58.04 & 30.70 & 15.27 & 38.64 \\
google/siglip-large-patch16-256 & 40.19 & 22.39 & 54.09 & 9.13 & 43.40 & 50.79 & 55.45 & 56.03 & 49.81 & 12.38 & 39.37 \\
laion/CLIP-ViT-H-14-laion2B-s32B-b79K & 33.03 & 19.14 & 58.82 & 21.81 & 54.09 & 60.23 & 52.92 & 55.50 & 33.11 & 15.41 & 40.41 \\
laion/CLIP-ViT-bigG-14-laion2B-39B-b160k & 38.84 & 20.44 & 60.90 & 25.02 & 55.42 & 59.95 & 62.27 & 57.86 & 35.02 & 16.21 & 43.19 \\
google/siglip-so400m-patch14-384 & 50.21 & 31.28 & 69.73 & 25.04 & 67.78 & 73.52 & 75.35 & 83.10 & 60.29 & 27.52 & 56.38 \\
\midrule
\multicolumn{12}{c}{\textbf{\textit{Ablation}}} \\
google/siglip-so400m-patch14-384 & 50.21 & 31.28 & 69.73 & 25.04 & 67.78 & 73.52 & 75.35 & 83.10 & 60.29 & 27.52 & 56.38 \\
google/siglip-so400m-patch14-384 (with vision-only paradigm) & 20.08 & 12.05 & 34.70  & 8.96 & 34.54 & 22.36 & 26.50 & 30.09 & 34.67 & 11.81 & 23.58 \\
$\Delta$ Performance Difference & \phantom{-}30.13 \textcolor{cherry}{\rlap{$\downarrow$}} & 19.23\textcolor{cherry}{\rlap{$\downarrow$}} & 35.03\textcolor{cherry}{\rlap{$\downarrow$}} & 16.08\textcolor{cherry}{\rlap{$\downarrow$}} & 33.24\textcolor{cherry}{\rlap{$\downarrow$}} & 51.16\textcolor{cherry}{\rlap{$\downarrow$}} & 48.85\textcolor{cherry}{\rlap{$\downarrow$}} & 53.01\textcolor{cherry}{\rlap{$\downarrow$}} & 25.62\textcolor{cherry}{\rlap{$\downarrow$}} & 15.71\textcolor{cherry}{\rlap{$\downarrow$}} & 32.80\textcolor{cherry}{\rlap{$\downarrow$}}\\
\midrule
\multicolumn{12}{c}{\textbf{\textit{Backbone}}} \\
Qwen2.5-VIT & 0.76 & 0.94 & 0.74 & 0.93 & 0.89 & 0.00 & 2.15 & 1.13 & 7.15 & 2.2 & 1.69\\
\multicolumn{12}{c}{\textbf{\textit{Ours}}} \\
\ours (mid-training only) & 35.81 & 20.46 & 67.61 & 37.20 & 75.09 & 66.30 & 68.02 & 67.00 & 63.96 & 27.90 & 52.94 \\ 
\ours (mid-training + finetuned) & 29.87 & 20.91 & 69.37 & 41.60 & 72.94 & 73.36 & 72.77 & 70.88 & 71.46 & 29.36 & 55.25\\
\bottomrule
\end{tabular}%
}
\caption{\ours Encoder performance on Visual Document Retrieval (VDR) Tasks, using ViDoRe benchmark, compared with SOTA baseline encoder models.}
\label{appendix:tab:VDR}
\end{table*}

\begin{table*}[ht]
\centering
\resizebox{\textwidth}{!}{%
\begin{tabular}{lccccccccccc}
\toprule
\textbf{Model name} & \textbf{ko-ko} & \textbf{ar-ar} & \textbf{en-ar} & \textbf{en-de} & \textbf{en-tr} & \textbf{es-en} & \textbf{es-es} & \textbf{fr-en} & \textbf{it-en} & \textbf{nl-en} & \textbf{Avg.} \\
\midrule
openai/clip-vit-base-patch32 & 18.10 & 28.30 & 8.25 & 22.15 & 17.97 & 12.15 & 47.56 & 19.48 & 22.74 & 25.05 & 22.18 \\
laion/CLIP-ViT-L-14-laion2B-s32B-b82K & 18.23 & 20.71 & 4.66 & 19.38 & 0.88 & 19.49 & 61.89 & 31.63 & 27.75 & 18.38 & 22.30 \\
laion/CLIP-ViT-B-32-laion2b-s34B-b79K & 16.25 & 21.73 & 4.20 & 17.82 & 17.37 & 25.07 & 57.03 & 22.91 & 21.49 & 23.38 & 22.72 \\
laion/CLIP-ViT-B-16-DataComp.XL-s13B-b90K & 19.21 & 18.40 & -1.69 & 33.07 & 6.57 & 16.93 & 62.39 & 20.93 & 19.40 & 32.23 & 22.74 \\
EVA02-CLIP-L-14 & 14.77 & 29.65 & 18.89 & 3.52 & 16.61 & 12.23 & 45.55 & 32.61 & 30.84 & 23.63 & 22.83 \\
EVA02-CLIP-bigE-14-plus & 11.36 & 31.51 & 10.71 & 24.33 & -10.05 & 20.18 & 59.20 & 36.12 & 28.60 & 33.18 & 24.52 \\
laion/CLIP-ViT-g-14-laion2B-s34B-b88K & 17.17 & 29.93 & 14.27 & 28.50 & -4.79 & 34.19 & 66.07 & 29.70 & 29.02 & 21.18 & 26.52 \\
EVA02-CLIP-bigE-14 & 10.97 & 29.99 & 13.49 & 22.76 & 6.39 & 29.03 & 57.16 & 36.66 & 33.43 & 26.16 & 26.60 \\
google/siglip-base-patch16-224 & 21.00 & 25.03 & 14.36 & 31.20 & 24.80 & 29.32 & 69.85 & 35.70 & 27.46 & 28.98 & 30.77 \\
google/siglip-base-patch16-256 & 21.40 & 30.46 & 12.67 & 30.19 & 19.81 & 28.50 & 71.68 & 36.55 & 28.75 & 30.72 & 31.07 \\
laion/CLIP-ViT-bigG-14-laion2B-39B-b160k & 14.38 & 32.39 & 12.21 & 36.74 & 14.99 & 30.44 & 69.77 & 39.77 & 36.44 & 34.83 & 32.20 \\
laion/CLIP-ViT-H-14-laion2B-s32B-b79K & 19.39 & 33.39 & 19.49 & 43.78 & 16.68 & 27.99 & 62.58 & 39.32 & 28.59 & 37.33 & 32.85 \\
openai/clip-vit-base-patch16 & 10.54 & 36.25 & 13.13 & 41.57 & 35.42 & 24.63 & 62.95 & 38.72 & 31.40 & 38.63 & 33.32 \\
laion/CLIP-ViT-L-14-DataComp.XL-s13B-b90K & 14.28 & 36.47 & 12.75 & 43.10 & 19.70 & 37.37 & 71.62 & 36.88 & 30.78 & 30.76 & 33.37 \\
openai/clip-vit-large-patch14 & 11.07 & 39.12 & 18.95 & 45.71 & 39.70 & 36.76 & 70.11 & 44.06 & 40.17 & 41.63 & 38.73 \\
google/siglip-so400m-patch14-384 & 13.65 & 45.76 & 11.22 & 46.07 & 30.62 & 40.08 & 73.62 & 46.36 & 36.45 & 44.95 & 38.88 \\
\midrule
\multicolumn{12}{l}{\textbf{\textit{Backbone}}} \\
Qwen2.5-VIT & 51.34 & 52.45 & 22.07 & 24.77 & 22.58 & 16.71 & 65.44 & 32.05 & 26.00 & 26.43 & 33.98\\
\multicolumn{12}{c}
{\textbf{\textit{Ours}}}\\
\multicolumn{12}{l}{\textbf{Mid-Training only}}\\
\ours & 51.13 & 50.96 & 2.09 & 66.00 & 54.59 & 55.51 & 70.66 & 63.00 & 61.91 & 62.46 & 53.83  \\ 
\ours (finetuned)& 49.95 & 43.85 & 6.96 & 63.40 & 50.98 & 58.19 & 76.38 & 62.17 & 62.22 & 62.73 & 53.68\\

{\textbf{Pretraining + Mid-Training}}\\

\ours & 57.51 & 50.13 & 1.44 & 67.4 & 55.86 & 61.42 & 75.82 & 67.73 & 67.88 & 66.36 & 57.16 \\
\ours (finetuned) & 57.24 & 51.61 & 3.59 & 68.55 & 49.70 & 64.23 & 80.48 & 67.11 & 64.19 & 65.39 & 57.21 \\
\bottomrule
\end{tabular}%
}
\caption{\ours Encoder Performance on Visual STS Tasks (Cross-lingual Tasks)}
\label{appendix:tab:sts_crosslingual_benchmarks}
\end{table*}

\begin{table*}[ht]
\centering
\resizebox{\textwidth}{!}{%
\begin{tabular}{lcccccccccc}
\toprule
\textbf{Model name} & \textbf{de} & \textbf{es} & \textbf{fr} & \textbf{it} & \textbf{nl} & \textbf{pl} & \textbf{pt} & \textbf{ru} & \textbf{zh} & \textbf{Avg.} \\
\midrule
openai/clip-vit-base-patch16 & 32.72 & 30.81 & 39.06 & 29.46 & 23.46 & 28.15 & 26.30 & 14.69 & 11.85 & 26.28 \\
EVA02-CLIP-B-16 & 30.68 & 27.02 & 36.05 & 27.13 & 29.71 & 32.41 & 29.06 & 25.40 & 16.71 & 28.24 \\
laion/CLIP-ViT-B-32-laion2b-s34B-b79K & 41.43 & 26.40 & 35.96 & 28.13 & 29.75 & 34.85 & 28.60 & 21.84 & 19.50 & 29.61 \\
laion/CLIP-ViT-L-14-laion2B-s32B-b82K & 39.99 & 31.22 & 40.69 & 28.57 & 28.49 & 27.58 & 25.85 & 22.66 & 22.58 & 29.74 \\
EVA02-CLIP-bigE-14 & 37.10 & 35.37 & 41.49 & 31.98 & 28.04 & 25.33 & 30.62 & 25.35 & 14.58 & 29.98 \\
laion/CLIP-ViT-B-32-DataComp.XL-s13B-b90K & 38.22 & 28.92 & 38.00 & 23.87 & 32.90 & 43.21 & 28.62 & 27.29 & 13.95 & 30.55 \\
openai/clip-vit-large-patch14 & 37.50 & 44.18 & 47.53 & 36.89 & 32.51 & 23.41 & 35.49 & 14.06 & 12.12 & 31.52 \\
EVA02-CLIP-bigE-14-plus & 31.96 & 37.53 & 46.88 & 38.94 & 29.78 & 27.50 & 33.35 & 25.05 & 16.20 & 31.91 \\
laion/CLIP-ViT-B-16-DataComp.XL-s13B-b90K & 41.25 & 31.76 & 45.92 & 34.60 & 35.79 & 40.38 & 36.57 & 26.67 & 15.18 & 34.24 \\
laion/CLIP-ViT-H-14-laion2B-s32B-b79K & 41.31 & 39.11 & 48.44 & 34.22 & 34.48 & 33.20 & 32.09 & 26.94 & 23.88 & 34.85 \\
google/siglip-base-patch16-224 & 40.38 & 41.80 & 45.75 & 37.90 & 37.64 & 42.65 & 37.01 & 32.81 & 10.79 & 36.30 \\
laion/CLIP-ViT-g-14-laion2B-s34B-b88K & 48.01 & 41.47 & 45.03 & 37.56 & 36.84 & 36.02 & 32.73 & 30.53 & 23.65 & 36.87 \\
laion/CLIP-ViT-bigG-14-laion2B-39B-b160k & 38.00 & 43.63 & 52.36 & 44.84 & 34.84 & 33.19 & 37.51 & 28.43 & 19.19 & 36.89 \\
google/siglip-base-patch16-256 & 42.40 & 44.36 & 46.72 & 41.73 & 38.72 & 42.34 & 39.56 & 35.01 & 9.34 & 37.80 \\
laion/CLIP-ViT-L-14-DataComp.XL-s13B-b90K & 47.05 & 45.13 & 50.76 & 44.24 & 38.21 & 34.94 & 37.87 & 30.89 & 14.65 & 38.19 \\
google/siglip-large-patch16-384 & 55.72 & 56.23 & 54.78 & 54.24 & 42.45 & 41.24 & 51.62 & 36.86 & 14.97 & 45.35 \\
\midrule
\multicolumn{11}{c}{\textbf{\textit{Backbone}}} \\
Qwen2.5-VIT  & 48.73 & 45.33 & 49.35 & 44.57 & 40.54 & 49.01 & 43.62 & 48.37 & 47.16 & 46.30\\
\multicolumn{11}{c}
{\textbf{\textit{Ours}}}\\
\multicolumn{11}{l}{\textbf{Mid-Training only}}\\

\ours & 64.56 & 61.74 & 67.21 & 63.24 & 60.14 & 59.38 & 58.90 & 60.78 & 60.19 & 61.79\\ 
\ours (finetuned) & 66.51 & 66.28 & 69.10 & 67.11 & 63.20 & 59.82 & 62.78 & 61.39 & 66.79 & 64.78\\

{\textbf{\textit{Pretraining + Mid-Training}}}\\

\ours & 66.94 & 66.03 & 70.07 & 66.73 & 65.12 & 64.34 & 62.97 & 62.18 & 63.04 & 65.27 \\
\ours (finetuned) & 68.56 & 70.18 & 72.21 & 69.95 & 66.51 & 62.84 & 67.91 & 64.76 & 68.27 & 67.91\\

\bottomrule
\end{tabular}%
}
\caption{\ours Encoder Performance on Visual STS Tasks (Multilingual Tasks)}
\label{appendix:tab:sts_multilingual_benchmarks}
\end{table*}

\end{document}